\documentclass{article}

\usepackage{microtype}
\usepackage{graphicx}
\usepackage{subfigure} % added
\usepackage{subcaption}
\usepackage{booktabs} % for professional tables

\usepackage{hyperref}

\usepackage[most]{tcolorbox}
\definecolor{color1}{HTML}{9ebcda}  % 浅雾蓝 
\definecolor{color2}{HTML}{B4DED5}  % 薄荷浅青蓝
\definecolor{color3}{HTML}{F7B186}  % 暖阳杏橙色
\definecolor{color4}{HTML}{bcbddc}  % 薰衣草浅紫色
\definecolor{color5}{HTML}{66c2a5}  % 嫩芽青绿色
\definecolor{color6}{HTML}{ff9999}  % 柔雾蜜桃粉色
\tcbuselibrary{listings}
\usepackage{listings}
\usepackage{pifont} 

\usepackage[accepted]{icml2026}

\usepackage{amsmath}
\usepackage{amssymb}
\usepackage{mathtools}
\usepackage{amsthm}

\usepackage{enumitem}
\usepackage[textsize=tiny]{todonotes}
\usepackage{etoc}
\usepackage{makecell}
\usepackage{wrapfig}

\usepackage[capitalize,noabbrev]{cleveref}

\theoremstyle{plain}

\theoremstyle{definition}

\theoremstyle{remark}

\usepackage[textsize=tiny]{todonotes}

\icmltitlerunning{Expectation Alignment of Language Models for Real-World User Expectations}

\begin{document}

\twocolumn[
  \icmltitle{Expectation Alignment of Language Models for Real-World User Expectations}
  
% Expecting Before Responding: Benchmarking and Latent-Guided LLMs
% From Expectations to Responses: 



  \icmlsetsymbol{equal}{*}

  \begin{icmlauthorlist}
    \icmlauthor{Miaomiao Li}{equal,cuhk,moe}
    \icmlauthor{Yang Wang}{equal,tecent}
    \icmlauthor{Bin Liang}{cuhk,moe}
    \icmlauthor{Shudong Liu}{macau}
    \icmlauthor{Zhiwei Zhang}{cuhk,moe}
    \icmlauthor{Kam-Fai Wong}{cuhk,moe}
  \end{icmlauthorlist}

  \icmlaffiliation{cuhk}{The Chinese University of Hong Kong}
  \icmlaffiliation{tecent}{Tencent Inc.}
  \icmlaffiliation{moe}{MoE Key Laboratory of High Confidence Software Technologies}
    \icmlaffiliation{macau}{University of Macau}

  \icmlcorrespondingauthor{Bin Liang}{bin.liang@cuhk.edu.hk}
  \icmlkeywords{Machine Learning, ICML}

  \vskip 0.3in
]



\printAffiliationsAndNotice{\icmlEqualContribution}

\begin{abstract}
Large language models (LLMs) have demonstrated remarkable performance on standard benchmarks, yet it remains largely unexplored whether they truly meet user expectations. 
Existing evaluation approaches, relying on model heuristics, expert rubrics, or user simulation, fail to capture the diversity and subtlety of real human expectations, causing models to appear competent while misaligning with what users actually seek. We present the first systematic study of user expectations in real-world LLM interactions, proposing a principled procedure to extract semantically rich expectations and introducing \textsc{ExpectBench}, a benchmark grounded in real user expectations. Analyses reveal that current LLMs struggle to satisfy and anticipate what users hope to obtain, highlighting a fundamental source of misalignment. 
Building on these observations, we propose \textsc{LENS}, a lightweight latent expectation–aware response generation framework. \textsc{LENS} enables models to internalize user expectations and generate better-aligned responses, consistently improving expectation satisfaction and underscoring the importance of explicitly modeling user expectations for realistic human–AI alignment. 
Code and data are available at \url{https://github.com/MiaomiaoLi2/expectation-alignment}. 
\end{abstract}

\section{Introduction}

\begin{figure}[t!]
  \centering
  \includegraphics[width=\linewidth]{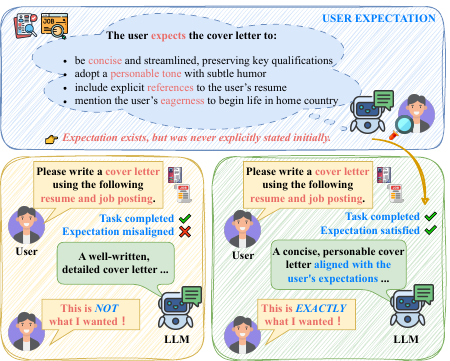}
  % \vspace{-.1in}
  \caption{When correct answers fail to meet user expectations.}
  % \vspace{-0.2in}
  \label{fig:user_expectation_gap}
\end{figure}

Large Language Models (LLMs) have achieved remarkable progress on a wide range of standardized benchmarks. They have surpassed most humans on exams such as the SAT~\citep{tan2024ai} and bar exam~\citep{katz2024gpt}, and even reached gold medal level on competitive programming~\citep{zou2025liveoibench} and mathematics competitions~\citep{huang2025gemini}. These successes have fostered a widespread impression that LLMs now perform at or beyond human level on many cognitively demanding tasks.
Yet, real-world usage tells a different story: a model’s response can be fluent, informative, and factually correct, and still fall systematically short of satisfying the user~\citep{mysore-etal-2025-prototypical, wang2024user}. 
As shown in Figure~\ref{fig:user_expectation_gap},
in interactive settings, success is not solely a matter of correctness; it depends on whether the response meets the diverse, often implicit expectations of individual users~\citep{wang2024understanding}.

The core challenge stems from the open-ended nature of human–AI interactions and the inherent multiplicity of user expectations. In real-world conversations, users’ goals are often underspecified~\citep{tamkintask}, expectations differ across individuals~\citep{kumar2024longlamp}, and there may be no single canonical answer~\citep{zheng2023judging}. For similar queries, different users may prefer distinct information, reasoning styles, or levels of detail, and even subtle contextual differences can induce a spectrum of expectations~\citep{wu2025personalized, lee2024aligning}. 
Critically, these expectations are rarely stated explicitly; instead, they are revealed implicitly through follow-up questions, clarifications, and corrections. 
Conventional evaluation paradigms, whether task-centric or capability-oriented, systematically overlook this diversity, focusing only on objective correctness or what an expert considers appropriate~\citep{mysore-etal-2025-prototypical, jin2025era}. 
As a result, even highly capable models can systematically fail to satisfy real users, exposing a fundamental gap between benchmark performance and practical utility~\citep{yao2025secondhalf}. Bridging this gap requires evaluation frameworks that capture diverse, multi-turn, and context-dependent user expectations, providing both a diagnostic lens and a foundation for guiding models toward expectation-aligned behavior. 

Prior work has attempted to evaluate open-ended interactions by employing LLMs as judges, using instance-specific checklists generated from the query \cite{linwildbench, pereira2024check}, or relying on domain experts to design unique rubric criteria \cite{arora2025healthbench, ruan2025expertlongbench, zhou2026general}. While these approaches provide consistency and scalability, the checklists and rubrics are based solely on model heuristics or expert judgment and do not necessarily reflect users’ expectations. As a result, evaluation reflects what experts or models consider appropriate rather than what users actually want.
Another line of work attempts to incorporate users more explicitly through user simulation~\citep{yao2025tau, qian2025userbench}. 
However, such simulations rely on simplified user models and constrained scenarios, limiting their ability to capture the full diversity and nuance of real human expectations.
Moreover, these evaluations typically focus on task completion or predefined metrics, rather than measuring whether models truly satisfy what users hope to achieve.

In this paper, we present the first systematic study of user expectations in real-world LLM interactions. We focus on three key questions: 
\begin{itemize}[leftmargin=1em,topsep=0pt]
\setlength\itemsep{0em}
\item \emph{RQ}1: How can user expectations be systematically extracted from multi-turn interactions, and how can they be organized into a benchmark for evaluating models? 
\item \emph{RQ}2: To what extent do existing LLMs satisfy these expectations, and what does this reveal about their alignment with real users? 
\item \emph{RQ}3: How can user expectations be explicitly modeled to guide response generation and improve user alignment?
\end{itemize}

For \emph{RQ}1, we analyze a large-scale dataset of 4.8 million multi-turn human–AI interactions in the wild, and design a principled procedure to extract user expectations from follow-up messages. These expectations are structured into semantically rich annotations that form the basis of a new benchmark, \textsc{ExpectBench}. It consists of 12,000 carefully curated interactions, with 6,000 examples each for training and testing, encompassing a total of 34,876 extracted user expectations. It covers diverse tasks, multiple languages, realistic multi-turn depths, and varied query lengths. To ensure stability and comprehensiveness, we leverage LLMs to summarize key expectation dimensions and adopt an iterative refinement process to consolidate and validate the resulting taxonomy.
% to ensure stability and comprehensiveness. 

For \emph{RQ}2, we evaluate a suite of representative LLMs on \textsc{ExpectBench}, measuring how well their outputs align with the extracted user expectations. All models exhibit relatively low alignment and substantial variance. Even the strongest model, GPT-4o, achieves an average score of only 2.72 out of 5, highlighting the difficulty of satisfying user expectations in real-world interactions. 
% To further understand the source of misalignment, we define a new task, Expectation Prediction, and examine whether models can anticipate user expectations from the initial query alone. 
To further understand the source of misalignment, we study \emph{Expectation Prediction} as a diagnostic evaluation setting, examining whether models can anticipate user expectations from the initial query alone.
On average, a human interaction reveals only 2.91 distinct expectations, yet even when models are allowed to predict up to 10 expectations per query, the best-performing model still achieves limited coverage, revealing that a key source of alignment failure is the model’s limited understanding of what users actually expect.  

Finally, for \emph{RQ}3, we introduce \textbf{\textsc{LENS}}, a two-stage lightweight \textbf{L}atent \textbf{E}xpectatio\textbf{N}--aware respon\textbf{S}e generation framework.  
\textsc{LENS} first enables the model to infer and internalize a latent representation of user expectations from the interaction context.
Conditioned on this latent signal, the frozen main LLM generates responses that better align with what users seek.
Experiments demonstrate that \textsc{LENS} consistently improves expectation satisfaction across dimensions and
models, highlighting the practical value of explicitly modeling user expectations.

This work makes three main contributions: 
1) Establishing user expectation alignment as a new evaluation paradigm for LLMs, 
with an explicit formulation of expectations expressed through follow-up interactions;
% in open-ended, multi-turn interactions, and proposing a principled method to extract semantically rich expectations from follow-up messages;
2) Introducing \textsc{ExpectBench}, the first large-scale benchmark grounded in real-world user expectations, revealing that even state-of-the-art LLMs struggle to satisfy and anticipate these expectations in practice;
3) Proposing \textsc{LENS}, a lightweight latent expectation–aware response generation framework 
% that consistently improves expectation satisfaction across dimensions and models.
% that enables LLMs to condition generation on inferred user expectations.
% that enables expectation-conditioned alignment without modifying the base LLM.
that conditions LLM responses on user expectation signals.
% , demonstrating the practical value of explicitly modeling user expectations.

\section{Related Work}

\begin{table*}[t]
\centering
\caption{
Comparison of user-centric LLM evaluation benchmarks.
``User-Focused'' indicates whether user intent, preference, or feedback is explicitly modeled as an evaluation target, rather than merely serving as a source of user queries.
``Evaluation Signal'' summarizes the primary form of feedback or criterion used to assess model behavior.
Prior user-centric benchmarks can be broadly categorized into three paradigms:
(i) using user data as query sources,
(ii) simulating users in controlled environments, and
(iii) estimating user satisfaction or behavior.
In contrast, our approach explicitly extracts and structures user expectations from real-world multi-turn interactions, enabling expectation-driven evaluation grounded in how users ultimately judge model outputs.
}
\small
\setlength{\tabcolsep}{5pt}
\begin{tabular}{lcccccc}
\toprule
\textbf{Benchmark} &
\makecell{\textbf{Real} \\ \textbf{User Logs}} &
\makecell{\textbf{Simulation-} \\ \textbf{Free}} &
\makecell{\textbf{User-} \\ \textbf{Focused}} &
\makecell{\textbf{Evaluation} \\ \textbf{Towards}} &
\makecell{\textbf{Evaluation} \\ \textbf{Signal}} \\
\midrule
WildBench~\cite{linwildbench} & 
\checkmark  & 
\checkmark & 
\texttimes & 
Model & 
Model-generated checklists \\

WildChat‑AQA~\cite{zhang-etal-2025-chat} & 
\checkmark & 
\checkmark & 
\texttimes & 
Model & 
Aggregative QA ranking \\

$\tau$-Bench~\cite{yao2025tau} & 
\texttimes & 
\texttimes & 
\checkmark &
Model & 
Final DB state \\

$\tau^{2}$-Bench~\cite{barres2025tau} &   
\texttimes & 
\texttimes & 
\checkmark & 
Model & 
Final DB state \\

IN3~\cite{qian2024tell} & 
\texttimes & 
\texttimes & 
\checkmark &
Model & 
Intent clarity \& task behavior \\

UserBench~\cite{qian2025userbench} & 
\texttimes & 
\texttimes & 
\checkmark &
Model & 
Preference elicitation rate \\

PENGUIN~\cite{wu2025personalized} & 
\texttimes & 
\checkmark & 
\checkmark &
Model & 
Personalized safety scores
 \\

SUPR~\cite{lin-etal-2024-interpretable} & 
\checkmark & 
\checkmark & 
\checkmark & 
User & 
Binary satisfaction patterns\\

RUBICON ~\cite{10.1145/3664646.3664778} & 
\checkmark & 
\checkmark & 
\checkmark & 
User & 
Binary satisfaction rubrics \\

\midrule
\textbf{\textsc{ExpectBench} (Ours)} & 
\checkmark & 
\checkmark & 
\checkmark & 
Model (User-defined) & 
Expectation Alignment\\
\bottomrule
\end{tabular}
% Our work is the first to evaluate LLMs against \emph{user expectations} explicitly extracted from real-world multi-turn interactions, treating users as the final arbiters of success.}
\label{tab:comparison}
\end{table*}

% \paragraph{User-Centric Evaluation}
Recent work has increasingly recognized the importance of evaluating LLMs from a user-centric perspective, moving beyond fully specified tasks toward realistic usage conditions.
A representative line of work collects \textbf{real-world user--LLM interaction logs}.
For example, WildChat~\cite{zhao2024wildchat} compiles large-scale conversations between human users and ChatGPT, providing a natural record of how users interact with LLMs in the wild.
Building on this data, WildBench~\cite{linwildbench} proposes an automated evaluation framework that benchmarks LLMs using challenging real-user queries.
However, its evaluation checklists are automatically generated by models directly from user queries, primarily reflecting the model’s interpretation of the task rather than the user’s post-hoc assessment.
% Moreover, over 70\% of WildBench instances are single-turn and filtered mainly by problem difficulty, favoring interactions that are easy in form but hard in content, rather than realistic usage settings in which user expectations are underspecified and only emerge through interaction and feedback.
Moreover, over 70\% of WildBench instances are single-turn and filtered by problem difficulty, favoring queries that are challenging in content but straightforward to answer, unlike everyday interactions where user expectations unfold through follow-up and refinement.

Another line of work evaluates LLMs through \textbf{simulated user interactions}.
% Benchmarks such as $\tau$-Bench~\cite{yao2025tau} evaluate language agents in interactive settings by assessing whether an interaction leads to the correct final database state, focusing on modular domains such as retail and airline services.
% Similarly, UserBench~\cite{qian2025userbench} simulates travel-planning scenarios in which users gradually reveal preferences in controlled environments.
Benchmarks such as $\tau$-Bench~\cite{yao2025tau} and UserBench~\cite{qian2025userbench} assess agents in controlled environments by measuring task completion or preference elicitation.
While effective for controlled experimentation, these benchmarks rely on simplified user models and narrowly scoped scenarios, limiting their ability to capture the richness and diversity of real user expectations.

More broadly, studies based on real user logs primarily focus on \textbf{measuring user experience or satisfaction}, often reducing feedback to coarse binary signals~\cite{lin-etal-2024-interpretable,10.1145/3664646.3664778,mahato-etal-2024-exploring}.
Recent analyses of WildChat further examine how users collaborate with LLMs across sessions, identifying prototypical behavioral patterns such as intent revision, exploration, and iterative refinement~\cite{mysore-etal-2025-prototypical}.
% While these works offer valuable insights into user behavior,
While valuable for understanding user behavior, their primary goal is to understand users rather than to leverage user feedback as a principled signal for model evaluation.

In contrast, our work treats users as the ultimate arbiters of success.
We explicitly extract and structure user expectations expressed through follow-up feedback, enabling expectation-driven evaluation grounded in how users actually judge model outputs in real-world interactions. 
To situate our work within the landscape of user-centric LLM evaluation, 
Table~\ref{tab:comparison} contrasts our benchmark with representative prior environments and benchmarks along several key dimensions.
% Table~\ref{tab:comparison} situates our benchmark among representative user-centric evaluation paradigms along several key dimensions.
Overall, existing benchmarks tend to emphasize user data, user simulation, or coarse user feedback in isolation, 
offering limited support for evaluating models against user expectations that unfold through interaction.

\section{\textsc{ExpectBench}: Benchmarking LLMs Against User Expectations}

\begin{figure}[t!]
  \centering
  \includegraphics[width=\linewidth]{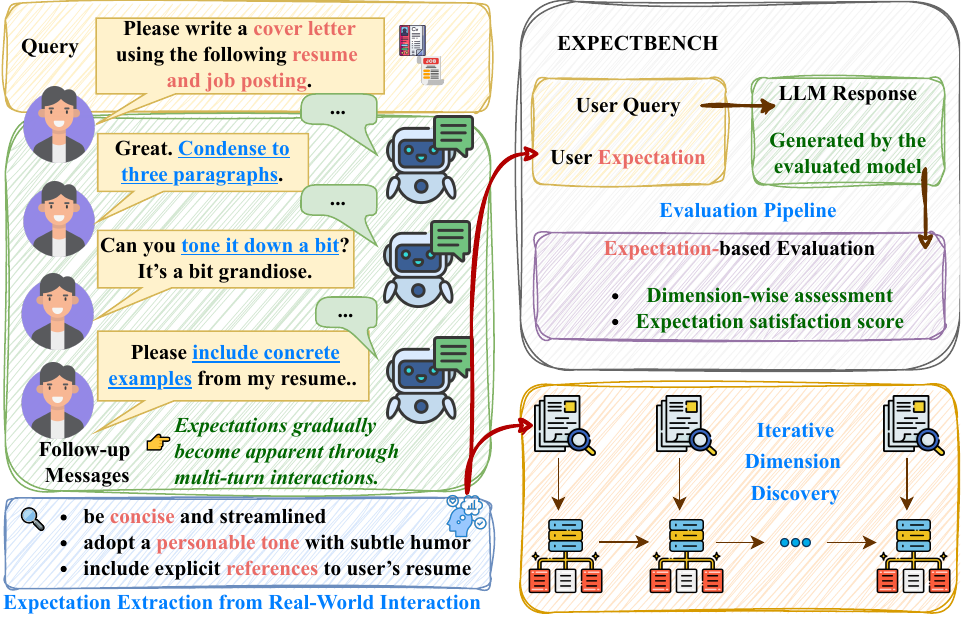}
  \caption{Overview of \textsc{ExpectBench}. 
    The left and bottom-right panels illustrate the processes of expectation extraction and iterative dimension discovery, 
    while the top-right panel depicts the expectation-based evaluation pipeline used to assess model responses in terms of expectation alignment.}
  \label{fig:fig_benchmark}
\end{figure}

In this section, we introduce \textsc{ExpectBench}, a benchmark designed to evaluate LLMs against user expectations inferred from real-world multi-turn interactions.
% Rather than relying on expert-defined rubrics or synthetic instructions, \textsc{ExpectBench} grounds evaluation criteria in user follow-up behavior, enabling systematic analysis of how well model responses align with what users actually seek.
As shown in Figure~\ref{fig:fig_benchmark}, \textsc{ExpectBench} is constructed by grounding evaluation criteria in user follow-up behavior, rather than expert-defined rubrics or synthetic instructions, enabling systematic analysis of how well model responses align with what users actually seek.

\subsection{Expectation Extraction}

User expectations are rarely stated explicitly in initial queries.
Instead, they are often revealed through follow-up turns, where users clarify requirements, correct assumptions, or express dissatisfaction with previous responses \cite{mysore-etal-2025-prototypical, jin2025era}.
We leverage this observation and extract expectations from multi-turn conversational trajectories rather than isolated queries.

Given a conversation consisting of an initial user query, a model response, and subsequent user turns, we treat follow-up messages as natural supervision signals that expose gaps between what the user expected and what the response provided.
Each expectation is formulated as a concise and self-contained criterion describing what the user hoped to obtain from the prior response (details in Appendix~\ref{app:expectation_extraction}).

To ensure meaningful evaluation, we retain only expectations that are clear and actionable, excluding purely conversational follow-ups (e.g., acknowledgments or topic changes) or those that are vague or ambiguous. 
This grounds \textsc{ExpectBench} in authentic user feedback, distinguishing it from benchmarks based on expert-designed rubrics or synthetic instruction templates.

\subsection{Benchmark Construction}

\textbf{Real-World Multi-Turn Conversations.}
\textsc{ExpectBench} is built from 4.8 million real human--AI interactions collected in the wild, based on the WildChat dataset~\citep{zhao2024wildchat}.
These conversations naturally capture how users refine requests, provide feedback, and clarify or redirect the interaction, making them well suited for studying user expectations.

\textbf{Data Filtering and Quality Control.}
To ensure high-quality, expectation-rich instances, 
% we apply filtering and verification procedures and human verification 
we apply automated filtering procedures followed by human verification for quality control (details in Appendix~\ref{app:data_filtering}).
The resulting conversations are coherent, informative, and suitable for evaluating expectation alignment in realistic interactions.

\textbf{Instance Construction. }
Each benchmark instance consists of a user query paired with expectations extracted from subsequent follow-ups.
These expectations serve as evaluation criteria reflecting what users actually emphasized or required, moving beyond task completion to capture practical usefulness.

\begin{figure}[h!]
  \centering
  \includegraphics[width=0.9\linewidth]{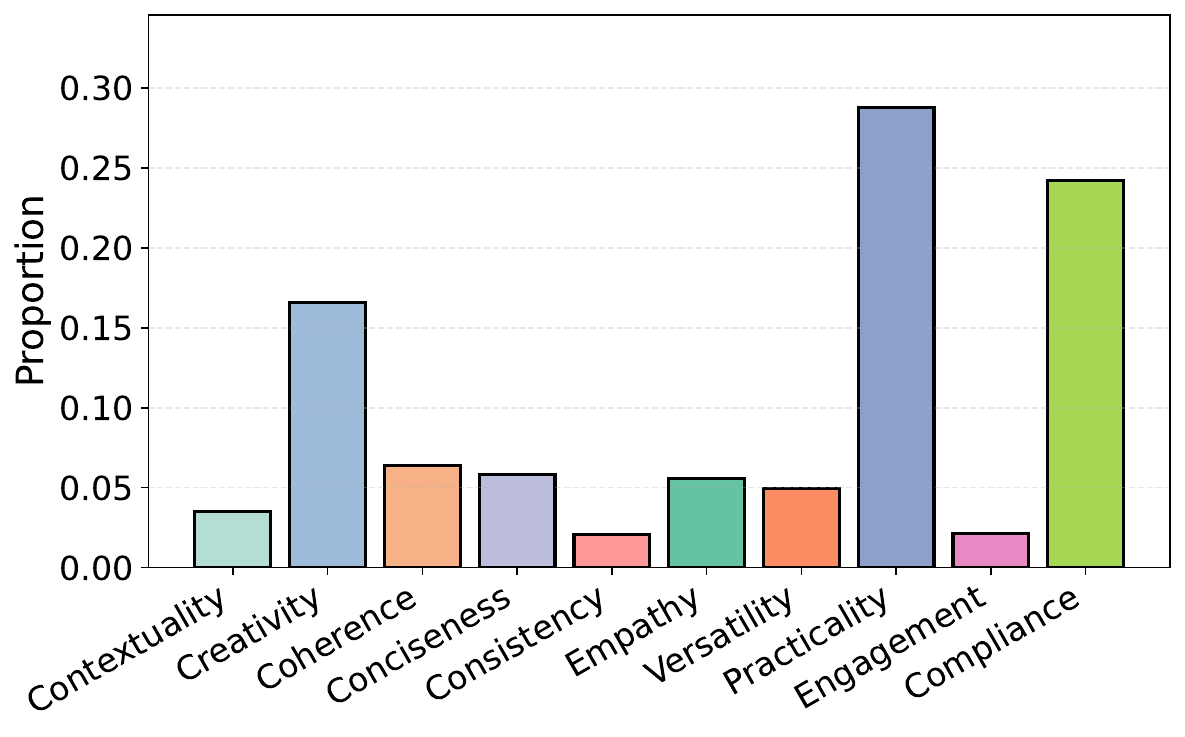}
  \caption{Expectation dimension distribution.}
  \label{fig:dimension_distribution}
\end{figure}

\begin{figure*}[!h]
	\centering
	\subfigure[Multi-turn Depth]{
	    \includegraphics[width=0.231\linewidth]{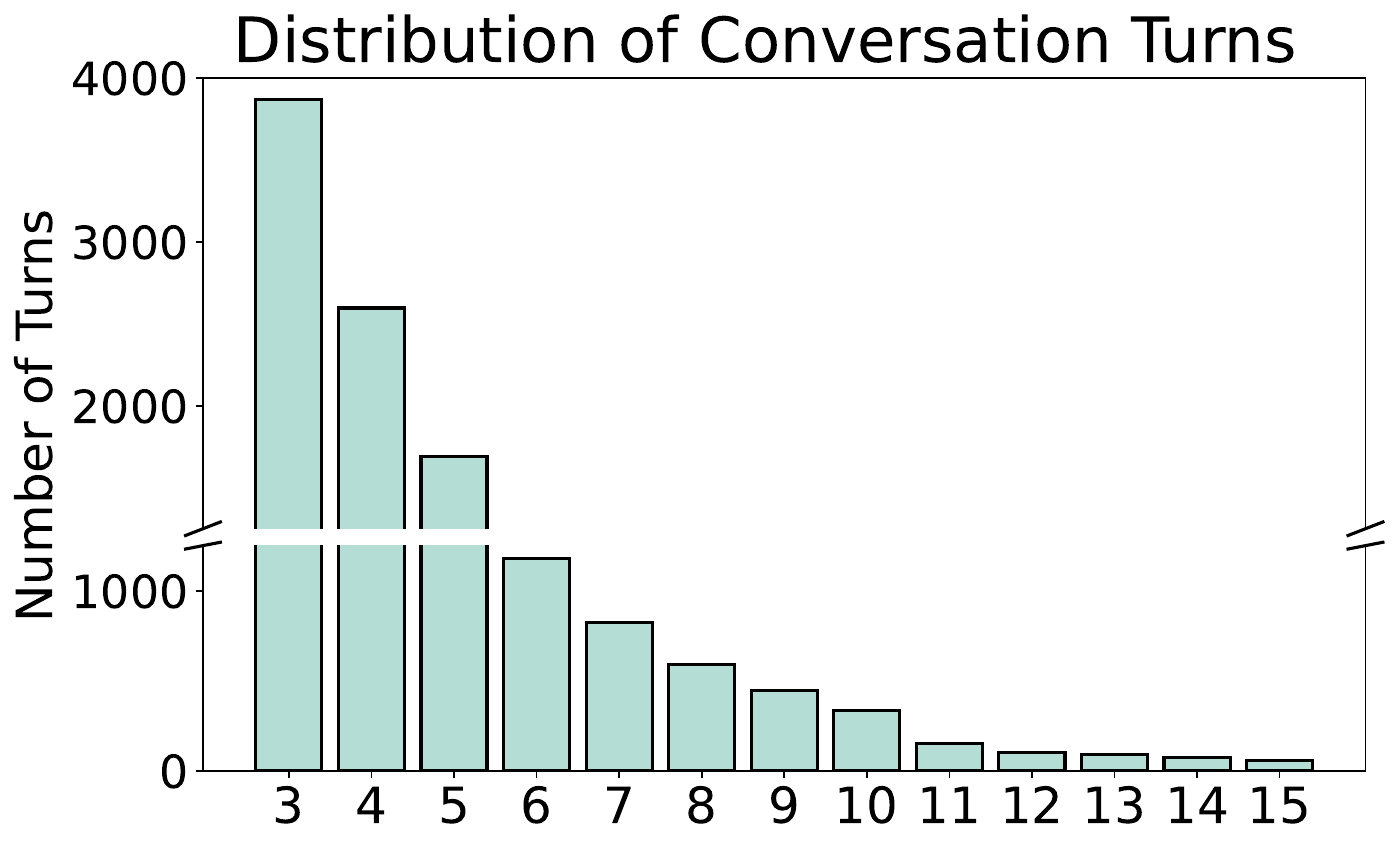}
	    \label{fig:statistic_turn}
        }
	\subfigure[Multilingual Coverage]{
	    \includegraphics[width=0.231\linewidth]{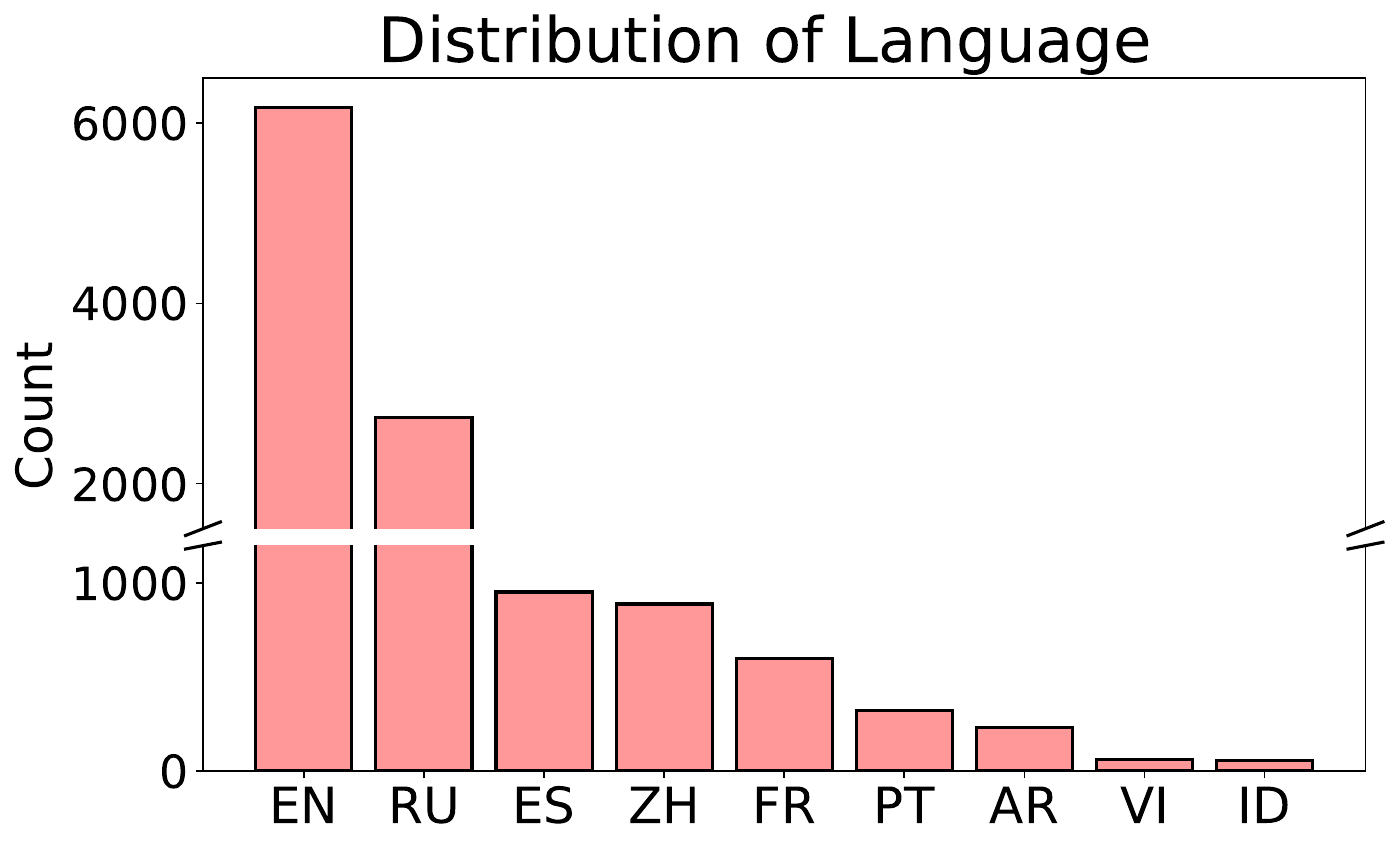}
	    \label{fig:statistic_language}
        }
	\subfigure[Expectation Richness]{
	    \includegraphics[width=0.231\linewidth]{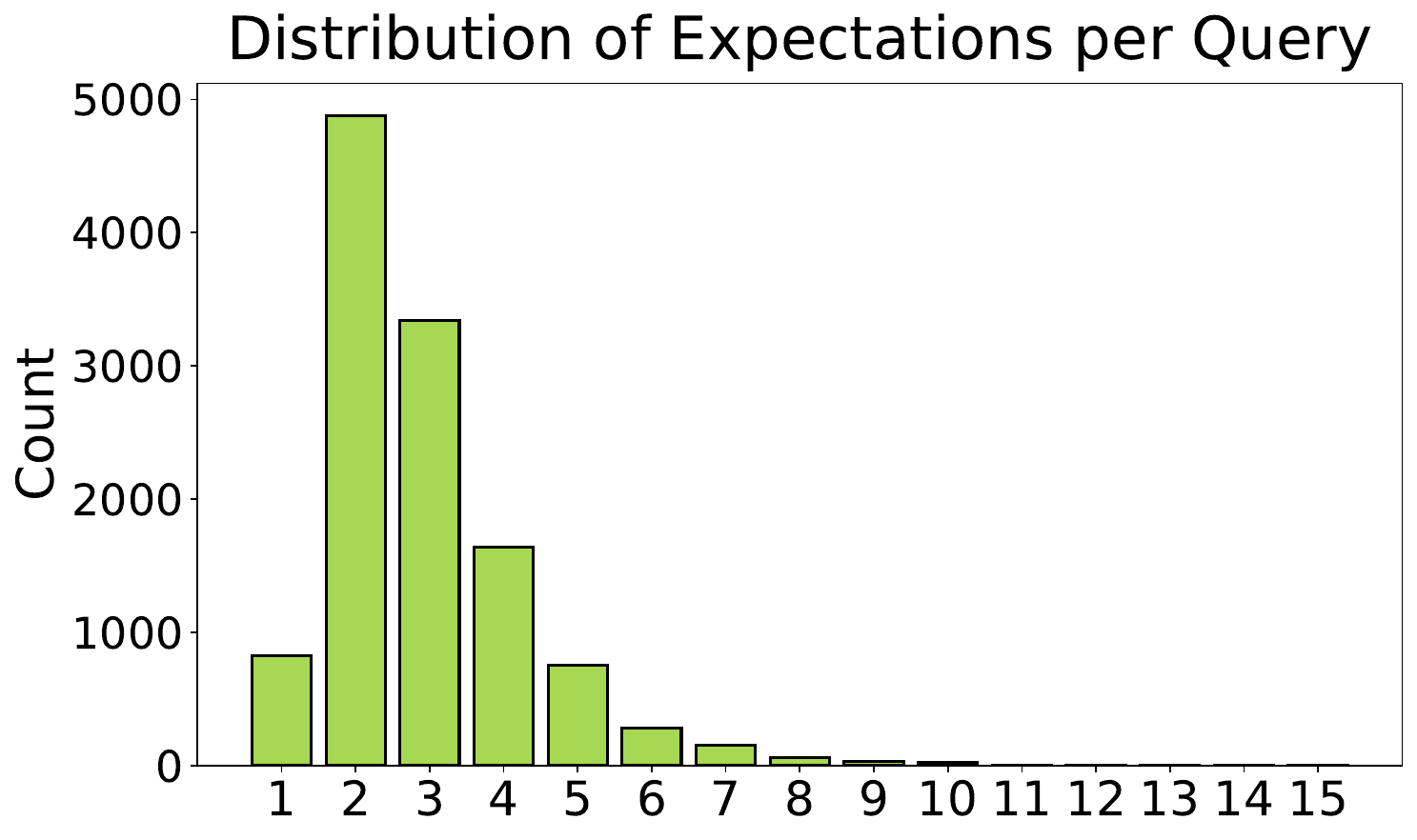}
	    \label{fig:statistic_expectation}
        }
	\subfigure[Realistic Query Lengths]{
	    \includegraphics[width=0.231\linewidth]{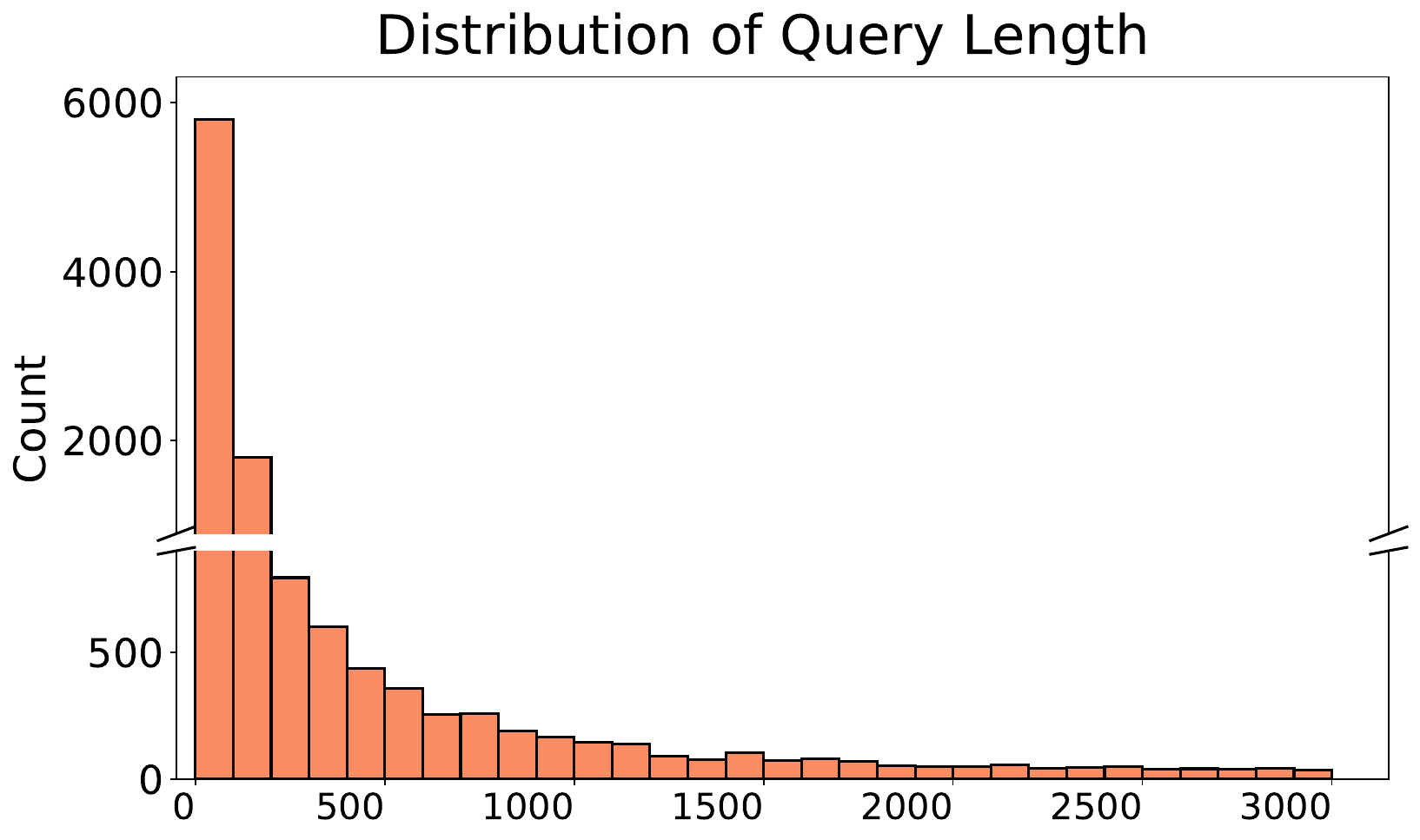}
	    \label{fig:statistic_length}
        }
        % \vspace{-.1in}
	\caption{Statistical overview of the \textsc{ExpectBench}.}
    % \vspace{-.1in}
          \label{fig:statistic}
\end{figure*}

\subsection{Dimension Discovery}

Given a collection of expectation annotations $\mathcal{E}$ extracted from user follow-up turns, our goal is to induce a compact set of semantically meaningful dimensions that systematically capture user expectations in real-world interactions. 
% Rather than relying on predefined taxonomies, we adopt a data-driven, iterative discovery process, drawing inspiration from prior work on interpretable satisfaction modeling~\citep{lin2024interpretable}.
Rather than relying on predefined taxonomies, we adopt a data-driven, iterative discovery process. 
This design is motivated by prior work on interpretable satisfaction modeling~\citep{lin2024interpretable} and scale-based AI evaluation~\citep{zhou2026general}, which suggests that decomposing evaluation into interpretable dimensions can provide more explanatory insight than a single aggregate score.

We randomly partition the full set $\mathcal{E}$ into sequential subsets $\{\mathcal{E}_i\}_{i=1}^n$, each containing a manageable number of annotations to respect context length constraints and ensure high-quality summarization. 
Let $\mathbf{Z}^{(i)}$ denote the dimension set after processing the $i$-th subset.
We iteratively update the dimension set as
% \[
$\mathbf{Z}^{(i)} = \mathrm{Refine}\big(\mathbf{Z}^{(i-1)}, \mathcal{E}_i\big),$
% \]
where $\mathrm{Refine}(\cdot)$ denotes the process of prompting a strong LLM to summarize newly observed expectations into a concise, interpretable set of evaluation dimensions, refining the set by merging, splitting, or rewriting dimensions as needed while keeping it compact and semantically meaningful.

After all subsets are incorporated, $\mathbf{Z}^{(n)}$ forms a stable taxonomy that captures diverse user expectations across functional, stylistic, and pragmatic aspects. 
Each expectation is assigned to its corresponding dimension, enabling multi-faceted evaluation of model outputs.

The resulting taxonomy spans dimensions including \emph{Contextuality}, \emph{Creativity}, \emph{Coherence}, \emph{Conciseness}, \emph{Consistency}, \emph{Empathy}, \emph{Versatility}, \emph{Practicality}, \emph{Engagement}, and \emph{Compliance}. 
Figure~\ref{fig:dimension_distribution} summarizes the empirical distribution of expectation annotations across the discovered dimensions, showing that a substantial fraction of user feedback emphasizes \emph{Practicality} and \emph{Compliance}, underscoring the importance of aligning LLM outputs with user intent rather than focusing solely on task correctness. 
We observe that many of these dimensions naturally align with recurring factors emphasized in prior studies on human-centered evaluation~\citep{wang2024user}, user experience~\citep{virvou2023artificial, xing2025factors}, and behavioral analysis~\citep{mysore-etal-2025-prototypical} of human--AI interaction.
Details of the discovery strategy, prompts, and 
the full taxonomy with definitions and representative examples are provided in Appendix~\ref{app:dimensions}.

\subsection{Benchmark Statistics}

\textsc{ExpectBench} comprises 12,000 interaction instances, each paired with user expectations extracted from subsequent follow-up messages. 
The dataset is split into training and test sets, supporting both expectation-aware model training and standardized evaluation of expectation understanding and alignment. In total, \textsc{ExpectBench} contains 34,876 instance-specific extracted user expectations.
Figure~\ref{fig:statistic} presents an overview of key benchmark statistics.

\textsc{ExpectBench} reflects realistic and diverse LLM usage scenarios. 
It spans nine languages, including English, Russian, Spanish, Chinese, French, Portuguese, Arabic, Vietnamese, and Indonesian, enabling evaluation of expectation understanding beyond English-centric interactions. 
All instances are derived from multi-turn conversations, where expectations emerge progressively through follow-up rather than being exhaustively stated in the initial query. 
On average, each instance contains 2.91 distinct expectations, highlighting the multi-dimensional nature of user feedback.

User queries are long and information-rich, averaging 366 tokens, distinguishing \textsc{ExpectBench} from datasets dominated by short prompts and better reflecting real-world tasks such as writing, planning, and complex instruction following. 
Detailed statistics on conversation length, language distribution, expectation cardinality, query length, and task categories are provided in Appendix~\ref{sec:benchmark_statistics}.

\subsection{Evaluation Metrics and Approach}

\textbf{Expectation-Based Scoring.}
\textsc{ExpectBench} evaluates model outputs against user expectations extracted from real-world multi-turn interactions.
For each instance, expectations revealed in user follow-up messages serve as explicit evaluation rubrics, specifying what the user required but did not obtain from the initial response.
We adopt a 5-point Likert scale (1: not satisfied at all; 5: fully satisfied) to measure the degree to which a model response aligns with each expectation.
When multiple expectations are associated with an instance, scores are first assigned at the expectation level and then aggregated to obtain an overall expectation alignment score, enabling both fine-grained and dimension-level analysis. 
Detailed evaluation prompts and data construction are provided in Appendix~\ref{sec:evaluation}.

\textbf{LLM-Based Evaluation and Reliability.}
EXPECTBENCH separates the evaluation target from the evaluator. 
For each instance, the target is defined by explicit expectation rubrics extracted from real user follow-up behavior, while the evaluator only assesses how well a response satisfies these given rubrics. 
This makes the benchmark compatible with different human or model-based judges without changing the underlying evaluation target.
Given a user query, model response, and the corresponding expectation rubrics, each response is scored on a 5-point scale based on explicit expectation alignment. 
While GPT-4o does not consistently perform well on expectation-aware generation tasks, it can serve as a reliable evaluator when explicit expectation information is provided~\cite{wu2025personalized,tanjudgebench,yuan2025s}. 
% In other words, difficulty in fulfilling expectations does not imply difficulty in assessing them.
% Given the large size of our evaluation set, fully relying on human annotations would be prohibitively expensive. 
To quantify its reliability, we conduct a consistency analysis across 283 expectation instances randomly sampled from our benchmark, comparing GPT-4o scores with four human annotations. The agreement is substantial, achieving a Cohen’s Kappa of $\kappa = 0.626$ and a Pearson correlation of $r = 0.963$ ($p < 0.001$).
These results indicate that, when conditioned on explicit expectations, GPT-4o can reliably assess expectation-level satisfaction. Based on this strong consistency, we adopt GPT-4o
as a scalable proxy for human evaluation in our experiments.
Detailed correlation statistics, human annotation guidelines, and examples are reported in Appendix~\ref{sec:evaluator_1} and \ref{sec:evaluator_2}.

\section{Results and Analysis}

Based on \textsc{ExpectBench}, we evaluate six representative LLMs that vary in model size,
training objectives, and intended capabilities.
The evaluated models include GPT-4o~\citep{hurst2024gpt}, 
DeepSeek-R1-7B~\citep{deepseekai2025},
GLM-4-9B~\citep{glm2024chatglm}, 
LLaMA-3.1-8B~\citep{grattafiori2024llama}, 
Mistral-7B~\citep{jiang2023mistral7b},
and Qwen3-8B~\citep{qwen3technicalreport}.
All models are evaluated under the same expectation-based scoring protocol.
% described in Appendix~\ref{sec:evaluation}.
Implementation details are provided in Appendix~\ref{sec:experiment_details}.

\subsection{Overall Benchmark Results}

\begin{figure}[t!]
  \centering
  \includegraphics[width=\linewidth]{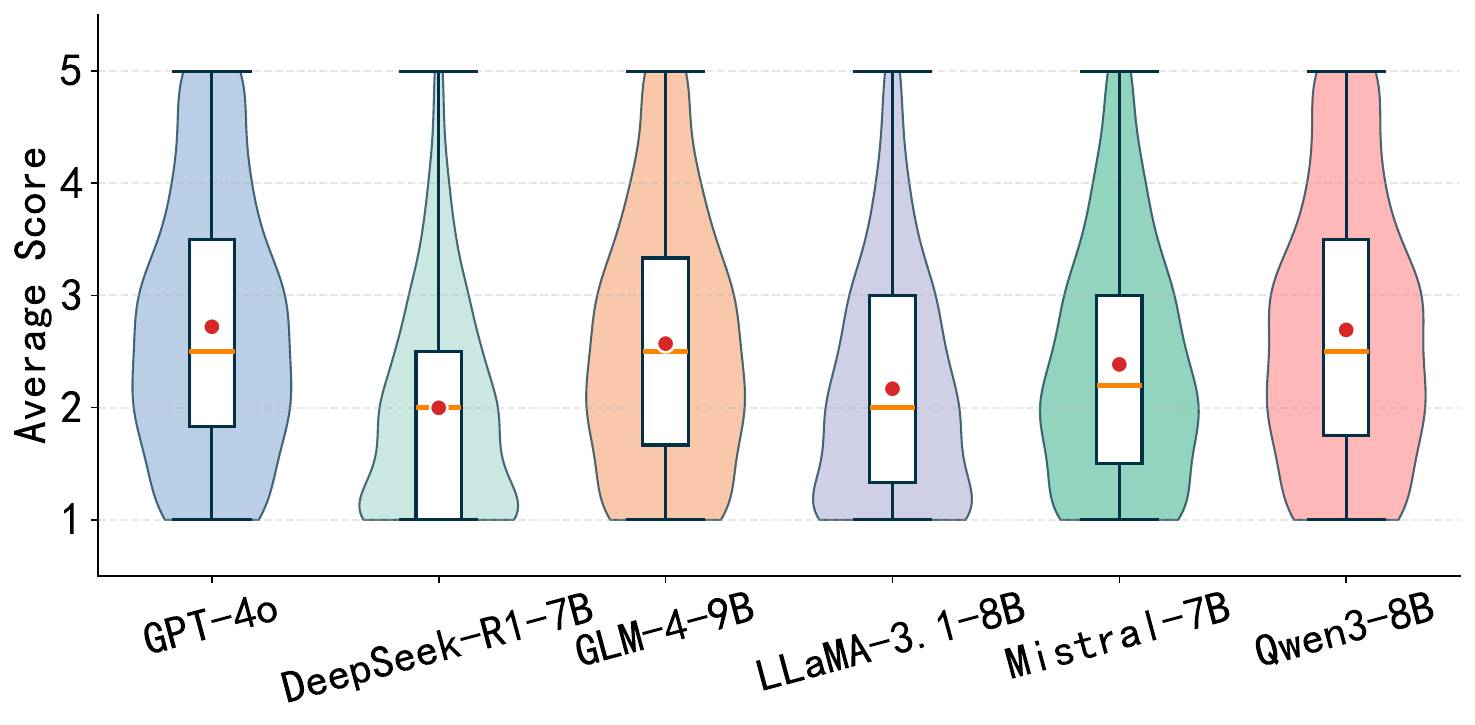}
  % \vspace{-.1in}
    \caption{
Distribution of instance-level expectation alignment scores across six LLMs.
% Each instance score is computed by averaging satisfaction scores over all associated user expectations on a 5-point Likert scale.
% The violin plot visualizes the full score distribution, with the central marker indicating the median.
Overall, all models exhibit relatively low expectation alignment and substantial variance, highlighting the challenge of satisfying user expectations that emerge from real-world multi-turn interactions.
}
% \vspace{-.15in}
  \label{fig:overall_results}
% \vspace{-0.1in}
\end{figure}

Figure~\ref{fig:overall_results} presents the overall expectation alignment performance of six representative LLMs
on \textsc{ExpectBench}.
For each test instance, we compute an instance-level score by averaging satisfaction scores across its associated user expectations.

Across all models, expectation alignment scores remain consistently low.
Even the strongest model, GPT-4o, achieves an average score of only 2.72 out of 5,
indicating that most responses only partially satisfy user expectations.
Open-weight models exhibit further degradation, with average scores ranging from 2.00 to 2.69.
These results suggest that current LLMs, regardless of scale or training paradigm, struggle to
fully meet user expectations expressed through real-world multi-turn interactions. 
Section~\ref{app:case} provides qualitative case studies that highlight common expectation–response mismatches.

Beyond mean performance, the distributional patterns reveal substantial variance across all models.
The wide interquartile ranges and long tails indicate unstable behavior, where models occasionally
achieve high expectation satisfaction but frequently fail to do so.
Notably, the performance gaps between models are smaller than the overall gap between model behavior
and full expectation satisfaction, underscoring a systemic limitation rather than model-specific failure.

Taken together, these findings demonstrate that satisfying explicit user expectations remains an open
and underexplored challenge for current LLMs, motivating the need for expectation-aware training and
evaluation frameworks.

\begin{figure}[t!]
  \centering
  \includegraphics[width=\linewidth]{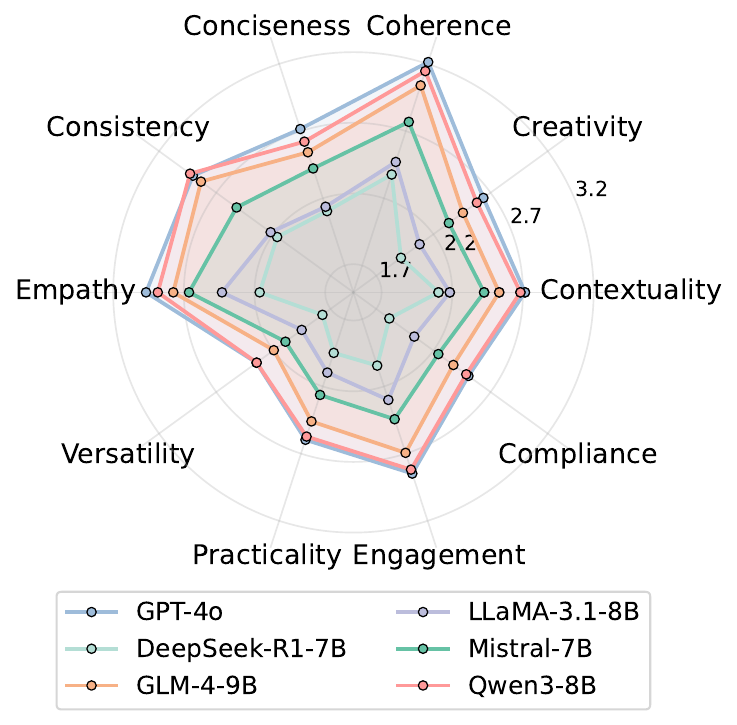}
    \caption{Expectation satisfaction across dimensions.
Average expectation alignment scores of different LLMs over ten expectation dimensions. No dimension achieves consistently high satisfaction, highlighting systematic challenges in meeting real-world user expectations.} 
% \vspace{-0.15in}
  \label{fig:dimension_results}
% \vspace{-0.1in}
\end{figure}

\subsection{Analysis across Expectation Dimensions}
\label{sec:dimension_results}

Figure~\ref{fig:dimension_results} presents model performance across the ten expectation dimensions defined in \textsc{ExpectBench}.
Across all evaluated models, expectation satisfaction remains consistently below the maximum score, indicating that no single dimension is reliably fulfilled in realistic user interactions.
% Figure~\ref{fig:dimension_results} presents model performance across the ten expectation dimensions in \textsc{ExpectBench}. 
% Overall, expectation satisfaction remains consistently below maximum scores, indicating that no single dimension is reliably fulfilled in realistic user interactions. 

Despite the overall low performance, clear dimension-level disparities emerge. 
Dimensions related to surface-level organization and presentation, such as \emph{Coherence} and \emph{Consistency}, achieve comparatively higher scores across models.
In contrast, dimensions that require deeper task understanding, real-world grounding, or strict adherence to user constraints, including \emph{Versatility}, \emph{Compliance}, and \emph{Practicality}, consistently receive the lowest scores.
This pattern is stable across model families, suggesting systematic challenges rather than model-specific weaknesses.

Further analysis of inter-dimension relationships and score distributions (see Appendix~\ref{sec:dimension_analysis}) shows weak correlations between dimensions and substantial variance within each. 
Frequent failures even in higher-scoring categories confirm that user expectations are inherently multi-dimensional and cannot be inferred from any single aspect alone.

\section{Predicting User Expectations from Queries}
\label{sec:expectation_prediction}

The consistently low expectation alignment scores observed  above suggest a fundamental limitation of current LLMs:
even when models generate fluent and seemingly reasonable responses, they
often fail to satisfy the criteria that users actually care about.
This raises a critical question beyond response quality itself:
\emph{do models understand what users expect in the first place?}

\begin{figure}[t!]
  \centering
  \includegraphics[width=\linewidth]{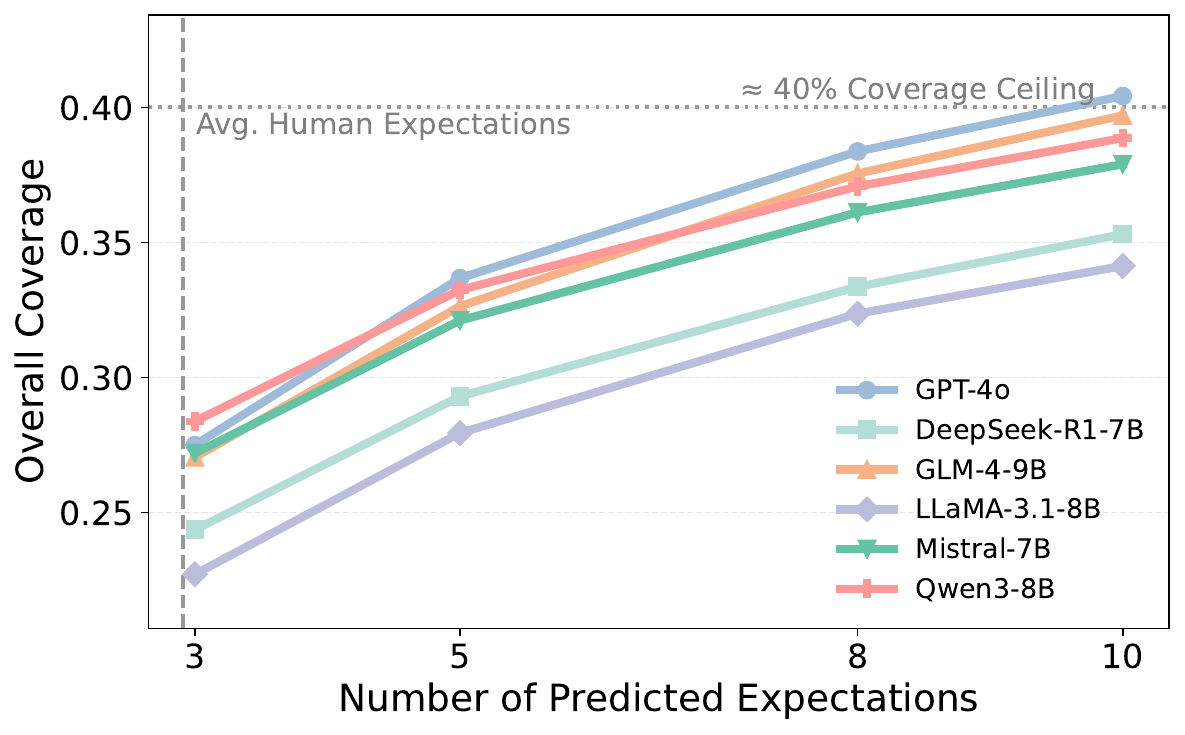}
  % \vspace{-.1in}
    \caption{Overall user expectation coverage at different prediction budgets.}
    % \vspace{-.1in}
  \label{fig:overall_coverage}
% \vspace{-0.1in}
\end{figure}
 
\subsection{Expectation Prediction and Metrics}
\label{sec:prediction_task_metric}

To investigate this question, we study expectation understanding as a
standalone capability.
Rather than evaluating whether a model can \emph{satisfy} expectations after seeing user feedback, we ask whether a model can \emph{anticipate} user expectations directly from the initial query, before any follow-up signals are revealed.
This setting isolates expectation inference from response generation and provides a diagnostic lens into the source of expectation misalignment.

Formally, given an initial user query $q$, a model is asked to predict a set of
expectations
\(
\hat{\mathcal{E}}_k(q) = \{\hat{e}_{q,1}, \dots, \hat{e}_{q,k}\},
\)
representing the top-$k$ aspects the model believes the user will rely on when
evaluating a response.
These predicted expectations are compared against the ground-truth expectation
set
\(
\mathcal{E}(q) = \{e_{q,1}, \dots, e_{q,m_q}\},
\)
which is extracted from follow-up user feedback.

We quantify expectation prediction performance using \emph{coverage}, defined
as the fraction of ground-truth expectations that are semantically matched by the model’s predictions.
Formally, coverage at prediction budget $k$ is computed as
\begin{equation}
\mathrm{Cov}_k
=
\frac{1}{N} \sum_{q=1}^{N}
\frac{
\big|\mathcal{E}(q) \cap \hat{\mathcal{E}}_k(q)\big|
}{
},
\end{equation}
where $N$ denotes the number of evaluation queries, and the intersection is determined by a semantic matching function that
identifies whether a predicted expectation meaningfully covers a ground-truth one.

We report coverage at multiple values of $k \in \{3, 5, 8, 10\}$, both overall
and per expectation dimension, providing a fine-grained view of prediction
performance. Additional details on semantic matching and
implementation are provided in Appendix~\ref{sec:appendix_prediction_details}.

\subsection{Results and Analysis}

Figure~\ref{fig:overall_coverage} presents overall expectation coverage as a function of the number of predicted expectations across six representative LLMs. 
Coverage grows with the number of predicted expectations, showing that models can partially approximate the space of user expectations given sufficient capacity. 
However, absolute coverage remains low: even when predicting far more expectations than the average 2.91 expressed by users, the strongest model achieves only around 40\%. 
This gap persists across model families and scales, suggesting that expectation prediction is intrinsically challenging rather than limited by prediction budget alone.

A finer-grained view is provided in Appendix~\ref{sec:predict_results}, which reports per-dimension coverage. 
Surface-level expectations related to organization and presentation are more likely to be anticipated, whereas dimensions requiring practical grounding, compliance awareness, or task-specific reasoning exhibit markedly lower coverage. 
These patterns mirror the dimension-level satisfaction trends, highlighting a strong correspondence between what models fail to anticipate and what they ultimately fail to satisfy.

Taken together, these results indicate that expectation misalignment in human--AI interaction arises primarily from models failing to infer user values, rather than from imperfect response generation alone. 
This motivates methods that explicitly model and leverage user expectations, instead of treating alignment as a byproduct of improved generation.

\begin{figure}[t!]
  \centering
  \includegraphics[width=\linewidth]{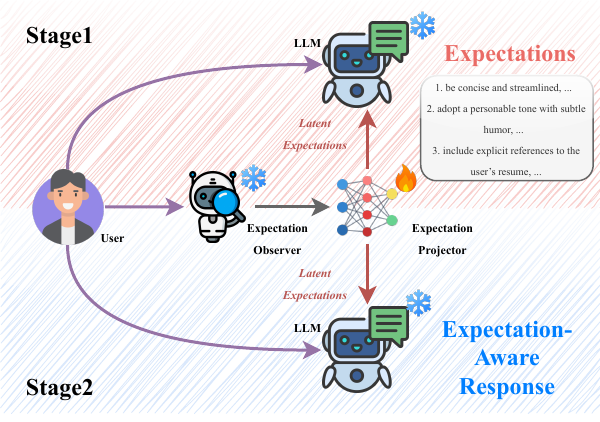}
  % \vspace{-.1in}
  \caption{The overview of our \textsc{LENS} framework.}
  % \vspace{-0.2i n}
  \label{fig:lens_framework}
\end{figure}

\section{\textsc{LENS}: Latent Expectation–Aware Response Generation}

\begin{table*}[t!]
\centering
\caption{Expectation satisfaction  on \textsc{ExpectBench} across ten dimensions and overall instances. 
Dimensions: Cx = Contextuality, Cr = Creativity, Co = Coherence, Cs = Conciseness, 
Ct = Consistency, Ep = Empathy, Vs = Versatility, Pr = Practicality, En = Engagement, Cm = Compliance.
\textbf{SFT-Exp} denotes models fine-tuned on expectation annotations only,
while \textbf{SFT-Resp} denotes models fine-tuned on GPT-4o generated responses that explicitly satisfy the extracted expectations.}
% \textbf{SFT-Exp} denotes models fine-tuned on expectation annotations only, whereas \textbf{SFT-Resp} denotes models fine-tuned on GPT-4o--generated responses that explicitly satisfy the extracted expectations.}
\label{tab:results_expectbench}
\begin{tabular}{r|cccccccccc|c}
\toprule
Model & Cx & Cr & Co & Cs & Ct & Ep & Vs & Pr & En & Cm & Overall \\
\midrule
Original (LLaMA-3.1-8B)  & 2.18 &2.08 &2.47 &2.14 &2.22 &2.43 &1.95 &2.10 &2.30 &2.03 & 2.17 \\
SFT-Exp (LLaMA-3.1-8B) & 2.21      &2.23      &2.57      &2.40      &2.38      &2.41      &1.96      &2.07      &2.32      &2.04  & 2.21    \\
SFT-Resp (LLaMA-3.1-8B) & 2.36      &\textbf{2.29}    &2.78      &2.46      &2.51      &2.53      &2.06      &2.21      &2.47      &2.14   & 2.33     \\
\textbf{\textsc{LENS} (LLaMA-3.1-8B)}   & \textbf{2.41} &\textbf{2.29} &\textbf{2.82} &\textbf{2.49} &\textbf{2.60} &\textbf{2.60} &\textbf{2.07} &\textbf{2.24} &\textbf{2.51} &\textbf{2.20} &  \textbf{2.37} \\
\midrule
Original (Mistral-7B)  & 2.42      &2.33      &2.77      &2.42      &2.52      &2.66      &2.09      &2.26      &2.44      &2.24  &  2.39  \\
SFT-Exp (Mistral-7B) & 2.39      &2.34      &2.77      &2.42      &2.56      &2.67      &2.06      &2.26      &2.48      &2.20    &  2.39 \\
SFT-Resp (Mistral-7B) & 2.41      &2.34      &2.89      &2.51      &2.56      &2.73      &2.12      &2.30      &2.45      &2.24    & 2.44\\
\textbf{\textsc{LENS} (Mistral-7B)}   &\textbf{2.52}      &\textbf{2.43}      &\textbf{2.92}      &\textbf{2.61}      &\textbf{2.65}      &\textbf{2.79}      &\textbf{2.20}      &\textbf{2.34}      &\textbf{2.57}      &\textbf{2.33}   &\textbf{2.50}  \\
\bottomrule
\end{tabular}
\end{table*}

Building on our analysis of expectation prediction, it becomes evident that a major bottleneck in aligning model outputs with user desires stems from the model’s limited understanding of the users’ expectations. 
Even state-of-the-art LLMs often fail to satisfy expectations because they do not internally represent what users value. 
This observation motivates the design of a framework that explicitly models latent user expectations and leverages them to guide response generation.

\subsection{Method}
\label{sec:method}

Directly fine-tuning large language models to satisfy user expectations is costly 
% can interfere with existing capabilities. 
and can easily affect their prior response behavior.
Explicit textual supervision is also suboptimal: user expectations are diverse, context-dependent, and often cannot be fully captured by prompts.

To address these challenges, we introduce  \textbf{\textsc{LENS}}, a  \textbf{L}atent \textbf{E}xpectatio\textbf{N}--aware respon\textbf{S}e generation framework. 
\textsc{LENS} consists of three roles.
An auxiliary model, termed the Expectation Observer, extracts latent representations of user expectations from the input query.
A lightweight trainable module, the Expectation Projector, transforms these representations into a form that is compatible with the main LLM.
The main LLM then generates responses conditioned on both the query and the projected latent expectations.

As shown in Figure~\ref{fig:lens_framework}, \textsc{LENS} operates in two stages.
In the first stage, the observer and projector enable the LLM to internalize expectation representations. In the second stage, the LLM produces expectation-aware responses, leveraging the latent signal without being distracted from the primary task. 
This two-stage design allows the model to internalize user expectations without requiring explicit textual supervision, while keeping the primary task objective intact.

\paragraph{Stage I: Expectation Observation and Internalization.} 
The first stage aims to endow the LLM with an internal understanding of user expectations. 
Given a query \( q \sim \mathcal{D}_{\text{train}} \) with annotated expectations \( \mathcal{E}(q) \),
the Observer \( \mathcal{O}_\phi \) takes \(q\) as input and produces an observer-side latent expectation representation \( \mathbf{z}(q) \):
\begin{equation}
\mathbf{z}(q) = \mathcal{O}_\phi(q).
\end{equation}
% The Projector \( \mathcal{P}_\psi \) then maps this representation into a projected representation \( \mathbf{h}(q) \) that is compatible with the main LLM:
The Projector \( \mathcal{P}_\psi \) then maps \( \mathbf{z}(q) \) into a projected expectation representation \( \mathbf{h}(q) \), aligned with the representation space of the main LLM:
\begin{equation}
\mathbf{h}(q) = \mathcal{P}_\psi(\mathbf{z}(q)).
\end{equation}
The main LLM, denoted as \(\mathcal{M}_\theta^\text{exp}\) when prompted for expectation prediction, is then conditioned on both \(q\) and \(\mathbf{h}(q)\) and predicts the expectation annotations:
% \[
\begin{equation}
\tilde{\mathcal{E}}(q) = \mathcal{M}_\theta^\text{exp}(\mathbf{h}(q), q).
% \] 
\end{equation}
We optimize the Projector parameters $\psi$ 
% by minimizing the expectation prediction loss
using the negative log-likelihood between the expectation annotations predicted by \(\mathcal{M}_\theta^{\text{exp}}\) and the gold annotations \(\mathcal{E}(q)\):
% \begin{equation}
% \mathcal{L}_{\text{exp}}(\psi) = \text{Loss}(\tilde{\mathcal{E}}(q), \mathcal{E}(q)),
% \end{equation}
% \begin{equation}
% \mathcal{L}_{\text{exp}}
% = - \log p_{\mathcal{M}_\theta^\text{exp}}
% \big( \mathcal{E}(q) \mid \mathbf{h}(q), q; \psi \big),
% \end{equation}
\begin{equation}
\mathcal{L}_{\text{exp}}(\psi)
= - \log p_{\mathcal{M}_\theta^{\text{exp}}}
\big( \mathcal{E}(q) \mid \mathbf{h}(q), q \big),
\end{equation}
% while keeping the Observer  \(\phi\) and LLM  \(\theta\) fixed. 
% while keeping the whole Observer $\mathcal{O}_\t h$ and LLM $\mathcal{M}_\theta^\text{exp}$ fixed. 
% while keeping the Observer $\mathcal{O}_\phi$ and  LLM $\mathcal{M}_\theta$ fixed.
while keeping the Observer parameters $\phi$ and the LLM parameters $\theta$ fixed.
Let $\psi^\ast$ denote the optimized Projector parameters after Stage I.
This trains the Projector to encode latent expectations in a form interpretable by the LLM without updating the Observer or the main LLM.

\paragraph{Stage II: Expectation-Aware Response Generation.} 
In the second stage, the trained Projector provides latent expectation guidance for response generation. 
% For a new query \(q\), 
Given a new query \( q' \sim \mathcal{D}_{\text{test}} \),
% we first obtain the latent representation:
the Observer first produces the latent signal \(\mathbf{z}(q')=\mathcal{O}_\phi(q')\), and the trained Projector maps it into the projected latent signal:
\begin{equation}
% \mathbf{h}^\ast(q') = \mathcal{P}_{\psi^\ast}(\mathcal{O}_\phi(q')),
\mathbf{h}^\ast(q') = \mathcal{P}_{\psi^\ast}(\mathbf{z}(q')).
\end{equation} 
% and the LLM produces the final response $r(q')$ conditioned on both the query and the latent signal:
% The main LLM then produces the final response $r(q')$ conditioned on both the query and the projected latent signal:
The same frozen main LLM, denoted as \(\mathcal{M}_\theta^\text{resp}\) when prompted for response generation, then produces the final response \(r(q')\) conditioned on both the query and the projected latent signal:
\begin{equation}
r(q') = \mathcal{M}_\theta^\text{resp}(q', \mathbf{h}^\ast(q')).
\end{equation}

In this stage, the latent expectation serves as soft guidance, enabling the LLM to generate responses aligned with user expectations while 
% maintaining its original task performance.
leaving the core model behavior unchanged.

\subsection{Results and Analysis}

We evaluate \textsc{LENS} on two base models, LLaMA-3.1-8B and Mistral-7B (details in Appendix~\ref{sec:method_details}). 
As shown in Table~\ref{tab:results_expectbench}, \textsc{LENS} consistently improves expectation satisfaction across both models, increasing overall scores from 2.17 to 2.37 for LLaMA and from 2.39 to 2.50 for Mistral.

% Beyond overall gains, improvements are particularly evident in both surface-level dimensions (e.g., Coherence and Consistency) and deeper expectation-oriented dimensions such as Empathy, 
Beyond overall gains, improvements are particularly evident in surface-level dimensions like \emph{Coherence} and \emph{Consistency}, as well as in deeper, expectation-oriented dimensions such as \emph{Empathy}, indicating that latent expectation guidance enhances not only response quality but also user alignment.
Notably, these improvements are achieved by training only a lightweight projector module, while keeping the base language models frozen.

A key observation from our results is that directly optimizing model parameters with stronger supervision, whether at the response level or the expectation level, does not reliably yield robust expectation alignment.
Response-level supervision primarily improves surface-consistent patterns (e.g., formatting or stylistic regularity), but fails to capture deeper, implicit expectations.
In contrast, expectation-level supervision provides highly instance-specific signals that vary substantially across users and contexts, making them difficult to generalize as a direct training objective.

By modeling expectations implicitly as latent guidance rather than explicit optimization targets, \textsc{LENS} avoids overfitting to subjective or noisy expectation signals while still enabling personalized alignment.
This design allows the model to leverage expectations as soft constraints instead of hard supervision, leading to consistent improvements across both shallow and deep dimensions, and demonstrating the importance of expectation-aware modeling for aligning language models with real-world user needs.
% By treating expectations as soft constraints rather than hard supervision, the model improves across both shallow and deep dimensions. This demonstrates the practical importance of expectation-aware modeling for aligning LLMs with real user needs.

\section{Conclusion and Future Directions}

In this work, we systematically investigated how well large language models align with real-world user expectations. 
Through the construction of \textsc{ExpectBench} and detailed analyses, we observed that even state-of-the-art LLMs often fail to satisfy and anticipate  what users actually seek, highlighting a fundamental source of misalignment in open-ended human–AI interactions. 
As a first step toward bridging this gap, we proposed \textsc{LENS}, a latent expectation–aware response generation framework that enables models to internalize user expectations and generate more aligned responses. 
The experiments demonstrate that \textsc{LENS} consistently improves expectation satisfaction across models and evaluation dimensions, underscoring the practical value of explicitly modeling user expectations.

Beyond these results, our study points to several directions for future research. First, investigating how models can infer and adapt to evolving expectations as multi-turn interactions unfold could improve dynamic alignment. Second, leveraging historical interaction to predict future expectations may enable more proactive and personalized responses. Third, exploring continual and dynamic learning approaches could allow models to adjust to shifting user preferences over time, supporting long-term expectation alignment.

We hope that our findings and resources, including \textsc{ExpectBench}, will serve as a foundation for developing LLMs that are not only capable but also truly responsive to the diverse and subtle expectations of real users.

\section*{Acknowledgements}
This work is partially supported by Hong Kong RGC GRF No. 14206324.
We would like to thank Zhipu AI for their support.
We also thank Prof. Jindong Wang from William \& Mary for his constructive comments.

\section*{Impact Statement}

This work provides the first systematic study of real-world user expectations grounded in multi-turn human–AI interactions, highlighting a critical gap between model outputs and what users actually value. By introducing \textsc{ExpectBench} and the \textsc{LENS} framework, we offer tools to better understand and improve expectation alignment in large language models. These contributions can guide the development of more user-centered AI systems that respond appropriately to diverse, context-dependent needs. 

Beyond immediate performance improvements, our findings emphasize the importance of explicitly modeling user expectations for safe and effective deployment of LLMs in real-world applications. While \textsc{LENS} represents a first step toward expectation-aware generation, it also points to broader challenges, including dynamic expectation tracking, personalized adaptation, and continual learning, which future work must address to ensure long-term alignment and responsible AI behavior.


\bibliography{expect}
\bibliographystyle{icml2026}

%%%%%%%%%%%%%%%%%%%%%%%%%%%%%%%%%%%%%%%%%%%%%%%%%%%%%%%%%%%%%%%%%%%%%%%%%%%%%%%
%%%%%%%%%%%%%%%%%%%%%%%%%%%%%%%%%%%%%%%%%%%%%%%%%%%%%%%%%%%%%%%%%%%%%%%%%%%%%%%
% APPENDIX
%%%%%%%%%%%%%%%%%%%%%%%%%%%%%%%%%%%%%%%%%%%%%%%%%%%%%%%%%%%%%%%%%%%%%%%%%%%%%%%
%%%%%%%%%%%%%%%%%%%%%%%%%%%%%%%%%%%%%%%%%%%%%%%%%%%%%%%%%%%%%%%%%%%%%%%%%%%%%%%
\newpage
\appendix
\onecolumn

\section{More Details of the \textsc{ExpectBench}} 

\subsection{Data Filtering and Quality Control} \label{app:data_filtering}
To ensure high-quality, expectation-rich evaluation instances, we apply a series of strict filtering and quality control procedures.

\textbf{Multi-turn focus.} We retain conversations with 3 to 15 turns, which are more likely to contain meaningful follow-up feedback while maintaining coherence and topical focus. Compared to prior benchmarks such as WildBench~\citep{linwildbench}, where over 70\% of instances are single-turn and filtered primarily by problem difficulty, \textsc{ExpectBench} emphasizes conversational depth and user follow-up behavior, which are essential for revealing implicit and explicit user expectations in real-world usage.

\textbf{Query and response filtering.} User queries shorter than 10 tokens or longer than 3,000 tokens are removed to exclude underspecified or unusually verbose prompts. Following prior work~\citep{jin2025era,pmlr-v267-yu25u}, we also require rejected responses to be at least 450 tokens long, using response length as a proxy for prompt quality. This ensures that the initial prompt elicits substantive model outputs, making subsequent user feedback more informative.

\textbf{Task-level filtering.} We annotate the task type of each conversation using a prompt classification tool\footnote{https://huggingface.co/valpy/prompt-classification}
 and remove instances labeled as \emph{coding} or \emph{mathematical reasoning or calculation}, which typically have well-defined correctness criteria and are less suitable for evaluating user expectations beyond task completion.

\textbf{Language and modality restrictions.} To ensure linguistic reliability, we limit the benchmark to the top 10 most frequent languages in the corpus and exclude multimodal interactions involving images, videos, or other non-textual content.

\textbf{Human verification.} Finally, we perform manual review to verify overall quality, remove noisy or degenerate cases, and ensure that selected instances are suitable for expectation-driven evaluation.

\subsection{More Details of Expectation Extraction}
\label{app:expectation_extraction}
\subsubsection{Prompt Template}

To extract user expectations from follow-up messages in multi-turn conversations, we prompt a strong LLM with a carefully designed instruction that emphasizes precision, semantic richness, and faithfulness to user-expressed intent. The prompt explicitly constrains the model to extract only expectations revealed in follow-up turns, avoiding restatement of the original request or speculative inference.

\begin{tcolorbox}[
  breakable,
    colback=color1!10!white, 
    colframe=color1!90!black,
  fonttitle=\bfseries,
  title=User Expectation Extraction Prompt
]

You are analyzing human--AI conversations to identify \textbf{user expectations}.

Given an initial user request and its follow-up messages, identify and extract the user’s meaningful expectations — the explicit or implicit preferences that indicate what the user wants the model’s response to achieve, convey, or emphasize with respect to the initial request.

When extracting expectations, ensure they satisfy the following criteria:

\begin{itemize}[leftmargin=1em]
  \item Do \textbf{not} restate the initial request.
  \item Do \textbf{not} represent a new or extended request that goes beyond the scope or subject of the initial user request.
  \item Do \textbf{not} include mechanical instructions without actionable guidance.
  \item Do \textbf{not} extract literal instructions, wording, surface-level content, or the multi-turn interaction process itself.
  \item Only extract expectations that are explicitly or implicitly expressed in the follow-up messages; do \textbf{not} infer unexpressed preferences.
  \item Only extract expectations that are semantically rich and meaningful, avoiding trivial or superficial ones.
  \item Extract multiple distinct expectations if present, each reflecting a unique aspect of the user’s intent.
  \item All expectations must be in English or translated into English.
\end{itemize}

\textbf{Output format (JSON):}

If no meaningful expectation is present:
\begin{verbatim}
{
  "has_expectation": "NO"
}
\end{verbatim}

If meaningful expectations are present:
\begin{verbatim}
{
  "has_expectation": "YES",
  "expectations": [
    {
      "content": "<Extracted expectation>"
    }
  ]
}
\end{verbatim}
\end{tcolorbox}

\subsubsection{Expectation Extraction Example}

This example illustrates how user expectations gradually become apparent through multi-turn interactions in a real-world human–AI conversation. The initial query specifies a task of writing a cover letter based on a resume and job posting, but many aspects of the desired output are left implicit. The follow-up messages progressively reveal nuanced and detailed expectations that the user had in mind from the beginning, providing insight into the user’s intent beyond what is explicitly stated.

The extracted expectations are highly detail-oriented. They specify not only the content to include or omit, such as removing personal contact information, but also structural preferences, for example condensing the cover letter to three paragraphs. These requirements often depend on interpreting cues implicitly embedded in the initial query, such as the scope of the task and the context provided by the resume and job posting.

Expectations are also diverse, covering multiple aspects of the response simultaneously. The user emphasizes the desired tone and style, calling for a personable and engaging voice with subtle humor, while also requesting that concrete details from the resume be incorporated to demonstrate relevant qualifications. This diversity reflects the multi-dimensional nature of user preferences, which go beyond simple correctness or completeness.

In addition, the expectations are highly personalized. Some reflect the user’s specific life circumstances, such as highlighting their recent return to America and eagerness to begin life in their home country. These expectations are tied to the individual user and would not be captured by generic writing guidelines. Many of these preferences are implicitly grounded, emerging naturally from the combination of the task, the resume, and the intended message rather than being explicitly spelled out in the initial request. For example, the desired length, the integration of personal narrative, and the cohesive flow of content all require inferring the user’s underlying goals and priorities.

Overall, this example demonstrates that user expectations are fine-grained, contextually grounded, and multi-dimensional. They combine detailed instructions, stylistic preferences, and personal information, highlighting the importance of evaluating LLM outputs not only against surface-level task requirements but also against rich, expectation-driven criteria derived from actual user intent.

\begin{tcolorbox}[
  breakable,
  colback=color6!5!white,
  colframe=color6!90!black,
  fonttitle=\bfseries,
  title=Example of Expectation Extraction
]

\textbf{User Query:}

\begin{verbatim}
Please write a cover letter using the following resume and job posting:
\end{verbatim}

\begin{tcolorbox}[
  breakable,
  colback=white,
  colframe=black!20,
  boxrule=0.5pt,
  listing only,
  listing options={
    basicstyle=\ttfamily\small,
    breaklines=true,
    breakatwhitespace=true,
    columns=fullflexible
  },
  title=Resume
]
\begin{verbatim}
John Moore
1815 Willow Oak Ln  
Sierra Vista, AZ 85635  
1-628-280-8919  
<PRESIDIO_ANONYMIZED_EMAIL_ADDRESS>  
makarah.net  
<PRESIDIO_ANONYMIZED_URL>
\end{verbatim} 
\vspace{0.1em} 
PROFILE 
\begin{verbatim}  
Multi-disciplinary Creative with 3+ years of experience in storytelling, 
video editing, and motion design, adept at creating engaging digital
content under tight deadlines. After 9 years in China, I am committed to 
bringing my versatile skillset to a high-energy US company promoting a
positive community culture.
\end{verbatim} 
\vspace{0.1em}
EDUCATION
\begin{verbatim}  
BA Film Studies, University of Colorado (2008)  
BFA Film Production, University of Colorado (2008)
\end{verbatim} 
\vspace{0.1em}
SKILLS
\begin{verbatim}  
• Video Editing
• 2D \& 3D Motion Design
• Copywriting
• Creative Writing
• Graphic Design
• Adobe Creative Suite
• Premiere Pro
• After Effects
• Illustrator
• Photoshop
• XD
• Blender 3D
\end{verbatim} 
\vspace{0.1em}
EXPERIENCE
\begin{verbatim}  
Game Writer & Content Creator, Mechanist Games (Oct 2021 -- Nov 2022)  
• Localized, proofread, and wrote marketing copy and in-game fiction.  
• Wrote, edited, and produced promotional videos for flagship titles.  
• Storyboarded advertising campaigns and created weekly comic content.  
• Collaborated with marketing and art teams across social media platforms.
\end{verbatim} 
\begin{verbatim}
Content Designer, 17Zuoye (Jan 2019 -- Mar 2021)  
• Created content for an AI-powered English learning app.  
• Wrote, directed, edited, and produced educational video series.  
• Collaborated with UI/UX teams on brand identity and app design.
\end{verbatim} 
\begin{verbatim}
English Teacher, Best Learning (Feb 2014 -- Dec 2019)  
• Prepared classes for 90+ students.
• Developed a new curriculum which fostered the transition between 
  preschool and elementary school.
• Adapted lessons and material to meet a wide range of children
  from 3 to 13 years old.
\end{verbatim} 
\end{tcolorbox}

\begin{tcolorbox}[
  breakable,
  colback=white,
  colframe=black!20,
  boxrule=0.5pt,
  listing only,
  listing options={
    basicstyle=\ttfamily\small,
    breaklines=true,
    breakatwhitespace=true,
    columns=fullflexible
  },
  title=Job Posting
]
\begin{verbatim}
Middle School Digital Media & Design Teacher  
Annie Wright Schools

Mission:
Cultivate individual learners to become well-educated, creative, 
and responsible citizens for a global society.

Role Overview:
The teacher designs and facilitates Digital Media and Design courses that 
integrate creative technologies, hands-on making, and human-centered 
problem solving. Projects follow the IB MYP Design Cycle and emphasize 
inquiry, iteration, and evaluation.

Key Requirements:
Proficiency with Adobe Creative Cloud.  
Experience or interest in interdisciplinary teaching.  
Strong communication skills and empathy for middle school learners.  
Eligibility to work in the United States.
\end{verbatim}
\end{tcolorbox}

\vspace{0.5em}

\textbf{Follow-up Messages:}

\begin{verbatim}
Follow-up 1:
Great. Condense to three paragraphs and remove my email address and phone
number.

Follow-up 2:
Hey, good job! Can you tone it down a bit? It's a bit grandiose.
Also try to add a bit more personality and fun!

Follow-up 3:
oops! This one was a bit too excited! Try again.
Maybe add a witty joke in there somewhere.

Follow-up 4:
OK, this is good. Can you add that I have just returned to America
from China and that I am eager to begin life in my home country?

Follow-up 5:
Oof, I don't like that cold opening. Can the opening be a bit more inviting? 
Overall, I like the direction on this one. Please try to make the China part
a bit more cohesive to the rest of the cover letter.

Follow-up 6:
Refrain from including mention of the International Baccalaureate framework.
Please include concrete examples from my resume to convince a recruiter of my 
qualifications. Condense to 3 paragraphs. Remember, add personality to it all.

Follow-up 7:
OK, let's go back to the following cover letter.
Just remove mention of the International Baccalaureate framework
and reduce it to 3 paragraphs.
\end{verbatim}

\vspace{0.5em}

\textbf{Extracted User Expectations:}

\begin{itemize}[leftmargin=1em]
  \item The user expects the cover letter to be concise and streamlined, preserving key qualifications while eliminating unnecessary details.
  \item The tone should be personable, engaging, and include subtle humor or elements of personality.
  \item Specific references to the user’s resume should be included to demonstrate relevant qualifications.
  \item The cover letter should mention the user’s return to America from China and their eagerness to start life in their home country.
  \item The narrative about returning from China should be integrated cohesively into the overall flow of the letter.
\end{itemize}
\end{tcolorbox}

\subsection{More Details of Dimension Discovery}
\label{app:dimensions}

\subsubsection{Dimension Discovery Procedure}

To construct the expectation taxonomy, we leverage GPT-4o to iteratively summarize the extracted expectation annotations. 
The full set of annotations is partitioned into 15 sequential subsets to accommodate context limitations and ensure high-quality summarization. 
For each subset, GPT-4o is prompted to integrate newly observed expectations into the current dimension set, optionally merging, splitting, or rewriting dimensions as needed to maintain interpretability and compactness. 
This iterative procedure yields a stable set of dimensions that captures the diversity of user expectations across functional, stylistic, emotional, and pragmatic aspects.

\subsubsection{Prompt Template}

To summarize user expectations into a concise set of evaluation dimensions, we use the following prompt. It guides the model to iteratively refine a set of abstract, user-centered dimensions suitable for evaluating AI responses.

\begin{tcolorbox}[
    breakable, 
    colback=color2!10!white, 
    colframe=color2!90!black, 
    fonttitle=\bfseries, 
    title=Dimension Discovery Prompt
]

\textbf{Task:} You are given a batch of user expectations extracted from human--AI conversations. Your task is to summarize these expectations into a concise set of \textbf{evaluation dimensions} that capture the underlying goals, preferences, or desired qualities users want from AI responses.

\textbf{Iterative Refinement:} You are performing a \textbf{multi-round refinement} of evaluation dimensions. In this round, you are provided with:
\begin{itemize}[leftmargin=1em]
    \item The \textbf{dimensions generated in the previous round} (if any).
    \item A \textbf{new batch of user expectations}.
\end{itemize}

\textbf{Requirements:}
\begin{itemize}[leftmargin=1em]
    \item Dimensions must be \textbf{abstract, user-centered, and intent-oriented}, reflecting what users ultimately want to achieve, not the literal wording of the expectations.
    \item Dimensions must be \textbf{rich, meaningful, and non-trivial} — capturing what would make an AI genuinely more useful to users.
    \item Dimensions must be \textbf{clear, concise, mutually exclusive, and non-overlapping}, suitable for evaluating AI responses.
    \item Use the \textbf{previous-round dimensions as a prior structure}:
    \begin{itemize}[leftmargin=1em]
        \item Preserve them when still supported by the current batch.
        \item Refine or merge them when there is partial overlap.
        \item Remove them when no longer meaningful.
        \item Create new dimensions only when necessary.
    \end{itemize}
    \item The goal is to \textbf{progressively converge} toward a stable global taxonomy of user-centered dimensions across batches.
    \item Limit the total number of dimensions to \textbf{8--12} for manageability. Only create as many as needed; fewer is acceptable if expectations cluster well.
    \item All output must be in English.
\end{itemize}

\textbf{Output format (JSON):}

Provide the dimensions as a numbered list in JSON format:

\begin{verbatim}
{
  "dimensions": [
    "Dimension 1: <short abstract name> – <descriptive explanation>",
    "Dimension 2: <short abstract name> – <descriptive explanation>",
    ...
  ]
}
\end{verbatim}
\end{tcolorbox}

\subsubsection{Detailed Descriptions and Examples}

In this appendix, we provide full descriptions of the expectation dimensions discovered from multi-turn user–LLM interactions. 
These dimensions were derived through an iterative, model-assisted summarization process followed by human verification.  
Each dimension captures a distinct aspect of user expectations and serves as a semantically rich criterion for evaluating whether model responses align with what users truly intend to obtain. 
Each dimension is listed with its full name, followed by the abbreviation used in the main text and figures in parentheses.

For each dimension, we also provide illustrative examples drawn from the benchmark. 
These examples span different levels of abstraction, ranging from high-level stylistic preferences to concrete content or formatting constraints, reflecting how user expectations naturally emerge in real-world interactions.

\textbf{Adaptability to Context and Task (Contextuality).}  
Reflects AI’s capability to adjust responses based on the user's specific context, task requirements, and preferences, ensuring relevance and personalization.
\begin{itemize}[leftmargin=2em, topsep=0pt]
    \item The user expects the conversation to include an anticipation of schedule changes in January.
    \item Ensure Cassy's written message to the viewer is consistent with her surroundings.
    \item Highlight Cassy's intrinsic appreciation for art and adaptation to the characteristics of different mediums.
\end{itemize}

\textbf{Creativity and Imagination (Creativity).}  
The ability of AI to craft inventive and novel responses that stimulate user engagement and thought.
\begin{itemize}[leftmargin=2em, topsep=0pt]
    \item Include a girl in a dreamy field with a small butterfly on her hand. 
    \item The user expects the document to be creatively written in a colloquial style, avoiding classical language to better engage with the reader.
    \item The user expects unique wording that differs substantially in style from standard output generated by chatgpt, including minor human-like errors for authenticity.
\end{itemize}

\textbf{Coherent Structuring and Clarity (Coherence).}  
Ensures AI’s proficiency in organizing content logically for clear understanding while maintaining a strong narrative flow.
\begin{itemize}[leftmargin=2em, topsep=0pt]
    \item The response should emphasize restructuring the names into a table format for clarity and organization.
    \item The user expects a more detailed discussion on how to reduce cyberbullying, extended to 800 words.
    \item Provide a strong final statement that ties the sources together, reviewing and analyzing their main points.
\end{itemize}

\textbf{Brevity and Precision (Conciseness).}  
AI’s skill in conveying information efficiently, using precise language without losing clarity or essential content.
\begin{itemize}[leftmargin=2em, topsep=0pt]
    \item Present the objectives in a more concise manner while maintaining clarity.
    \item The response should effectively condense the content to highlight the main points, focusing on essential information while maintaining clarity.
    \item Ensure that each section of the presentation lasts less than 10 minutes.
\end{itemize}

\textbf{Consistency and Professional Formatting (Consistency).}  
The AI's ability to deliver content with a uniform style and format that enhances readability and appears professional.
\begin{itemize}[leftmargin=2em, topsep=0pt]
    \item Maintain professional and clear slide presentation style to ensure easy readability.
    \item Use straight parentheses and quotes instead of curved ones to prevent formatting issues
    \item Make the interview script more structured and formal.
\end{itemize}

\textbf{Emotional Intelligence and Genuine Interaction (Empathy).}  
The capacity of AI to recognize, adapt, and respond appropriately to emotional nuances in human interactions, fostering empathetic and sincere communications.
\begin{itemize}[leftmargin=2em, topsep=0pt]
    \item The script should humorously highlight and mock translation errors and inconsistencies, providing comedic relief through the characters' playful and exaggerated reactions.
    \item The story should include the emotional impact of Tatiana naming their daughter Mona Rose, highlighting Butcher's sentimental reaction due to the personal significance of the name.
    \item Ensure the content tone is fun, engaging, and highly feminine, avoiding topics perceived as male-centric, while still being inclusive to men.
\end{itemize}

\textbf{Versatile Problem Solving (Versatility).}  
AI's aptitude in accommodating diverse user issues and objectives, offering multiple methods to effectively resolve specific challenges.
\begin{itemize}[leftmargin=2em, topsep=0pt]
    \item Identify what types of personnel are needed for the project team of the 'one-click publishing' theme.
    \item Consolidate similar processes within the diagram to streamline the workflow.
    \item Develop and formalize mathematical or computer models to optimize resource allocation in subway construction projects.
\end{itemize}

\textbf{Practical Insight and Grounded Perspective (Practicality).}  
AI’s proficiency in incorporating practical knowledge and contextual understanding to provide realistic and applicable insights.
\begin{itemize}[leftmargin=2em, topsep=0pt]
    \item Elaborate on how consumers can benefit from the reduced environmental impact.
    \item Formulate insightful questions for the fact-finding process that effectively aid in determining the appropriate insurance coverage.
    \item The user expects a reflection on the integration of artificial intelligence in mobile app development, particularly in educational curricula, and how this integration can help students understand practical applications of technology in solving complex problems.
\end{itemize}

\textbf{Engagement and Interactivity (Engagement).}  
The AI's competence in crafting responses that are engaging, stimulating user interaction, and maintaining interest throughout the communication.
\begin{itemize}[leftmargin=2em, topsep=0pt]
    \item Encourage engagement with practices of spiritual development and life enrichment related to tea meditation.
    \item The user expects a more detailed and engaging expansion on the characteristics of individuals with a Gemini Sun and Capricorn Moon, making the content longer and more captivating.
    \item Include examples and exercises in the lesson to enhance understanding.
\end{itemize}

\textbf{Accuracy and Compliance (Compliance).}  
AI’s accuracy in generating responses that align with factual correctness, adherence to guidelines, standards, or specific user-defined directives.
\begin{itemize}[leftmargin=2em, topsep=0pt]
    \item Ensure the revised poem is direct and retains the original writing style to reassure readers about the author's authenticity.
    \item The user expects clarification and correction on incorrect or non-functional script or method suggestions.
    \item Must provide necessary information for each fable according to the specified SEO model.
\end{itemize}

\subsection{Benchmark Statistics}
\label{sec:benchmark_statistics}

To provide an overview of the scale, diversity, and structural properties of \textsc{ExpectBench}, we present a statistical analysis of the benchmark dataset in Figure~\ref{fig:statistic}.

\textbf{Dataset Scale and Splits.}
\textsc{ExpectBench} contains a total of 12,000 multi-turn conversation instances, each paired with explicitly extracted user expectations derived from follow-up feedback.
We split the dataset into training and test sets, which are used respectively for expectation-aware model development and for evaluating both expectation prediction and expectation-aware response generation. In total, \textsc{ExpectBench} contains 34,876 extracted user expectations. 
This design enables the benchmark to support not only model comparison, but also supervised and fine-tuning-based approaches for improving expectation alignment.

\textbf{Conversation Turn Distribution.}
Figure~\ref{fig:statistic_turn} illustrates the distribution of conversation lengths (in turns) among selected instances.
All conversations contain multiple turns, consistent with our focus on expectation emergence through interaction.
The majority of instances fall within a moderate turn range, indicating that expectations are typically revealed through a small number of focused follow-up messages rather than prolonged dialogues.
This supports our core premise that expectation mismatches arise naturally in everyday interactions, without requiring complex or lengthy conversations.

\textbf{Multilingual coverage.}
As shown in Figure~\ref{fig:statistic_language}, the benchmark exhibits broad multilingual coverage.
The dataset spans nine major languages—English (EN), Russian (RU), Spanish (ES), Chinese (ZH), French (FR), Portuguese (PT), Arabic (AR), Vietnamese (VI), and Indonesian (ID)—reflecting the linguistic diversity of real-world LLM usage.
This multilingual setting allows us to evaluate whether expectation understanding and alignment generalize beyond English-centric interactions, and exposes additional challenges introduced by cross-lingual pragmatics and cultural variation.

\textbf{Expectation Cardinality.}
Figure~\ref{fig:statistic_expectation} reports the distribution of the number of expectations extracted per instance.
On average, each conversation contains 2.91 user expectations, confirming that user feedback is often multi-faceted.
Rather than correcting a single failure mode, users frequently comment on multiple aspects of a response—such as tone, structure, usefulness, or contextual appropriateness—highlighting the limitations of single-criterion evaluation and motivating expectation-aware assessment.

\textbf{Query Length Distribution.}
We further analyze the length of initial user queries in Figure~\ref{fig:statistic_length}.
The average query length is 365.94 tokens, reflecting the prevalence of complex, information-rich prompts in real-world usage (e.g., writing tasks, planning requests, or context-heavy instructions).
This distribution distinguishes \textsc{ExpectBench} from benchmarks dominated by short or synthetic prompts, and underscores the importance of evaluating models under realistic input conditions.

\begin{figure}[!h]
    \centering
    \includegraphics[width=0.5\linewidth]{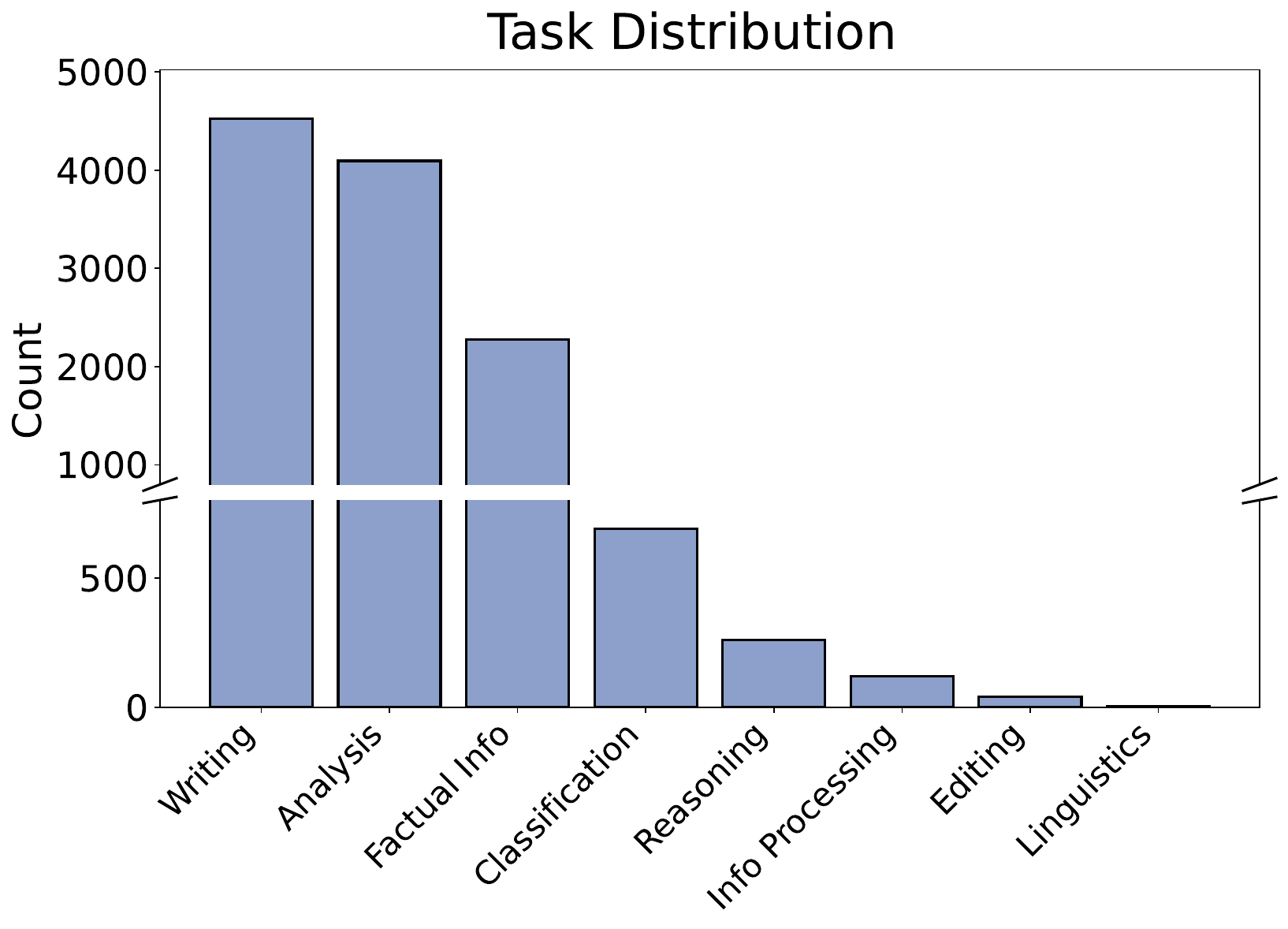}
    \caption{Distribution of task categories in \textsc{ExpectBench}.}
    \label{fig:statistic_task}
\end{figure}

\textbf{Task Diversity.}
The benchmark covers a wide range of open-ended tasks, including writing, explanation, planning, advice-seeking, and creative generation.
Figure~\ref{fig:statistic_task} presents the distribution of task categories across the benchmark.
By focusing on tasks where expectations extend beyond factual correctness, the benchmark emphasizes scenarios in which user satisfaction depends on nuanced alignment rather than objective answers.

Overall, these statistics demonstrate that \textsc{ExpectBench} captures diverse, multilingual, and expectation-rich interactions representative of real-world LLM use, providing a robust foundation for evaluating how well models anticipate, interpret, and satisfy user expectations.

\subsection{Evaluation Metrics} \label{sec:evaluation}

\subsubsection{Evaluation Prompt}

To rigorously assess how well AI responses align with user expectations, we use the following prompt. It guides the model to perform strict, evidence-based scoring for each expectation extracted from human–AI conversations.

\begin{tcolorbox}[breakable, colback=color3!10!white, colframe=color3!90!black, fonttitle=\bfseries, title=Expectation Alignment Evaluation Prompt]

You are a strict and detail-oriented evaluator.  
Your role is to assess expectation alignment between a user’s intent and an AI-generated response.  
You must apply conservative judgment and evaluate based on explicit evidence only. Generous or assumption-based scoring is strictly prohibited.

You will be given:
\begin{itemize}[leftmargin=2em]
    \item A user query
    \item An AI-generated answer
    \item A list of user expectations extracted from the conversation
\end{itemize}

For EACH expectation, determine whether the answer satisfies the exact wording, scope, and all required details of the expectation.

Important Evaluation Principles:
\begin{itemize}[leftmargin=1em]
    \item Each expectation must be evaluated independently.
    \item An expectation is considered satisfied ONLY if the answer explicitly and directly addresses its requirement.
    \item Vague, generic, or loosely related content must NOT be considered sufficient.
    \item If an expectation contains multiple required elements, ALL elements must be satisfied. Missing any required element constitutes a failure to satisfy the expectation.
    \item Do NOT infer satisfaction based on intent, tone, or plausibility; rely only on explicit evidence in the answer.
\end{itemize}

Scoring rules (Expectation-Level):
\begin{itemize}[leftmargin=1em]
    \item 5 — Fully satisfied:  
    All required elements of the expectation are explicitly and clearly fulfilled.  
    The response directly addresses the expectation’s core requirement.  
    No critical details are missing.  
    Evidence is specific and unambiguous.

    \item 4 — Mostly satisfied:  
    All required elements are present.  
    Minor weaknesses exist (e.g., limited detail, slight lack of clarity).  
    No required element is missing.  
    Use this score only when all elements are covered.

    \item 3 — Partially satisfied:  
    Only some required elements are fulfilled.  
    One or more key components of the expectation are missing.  
    The response addresses the expectation incompletely.

    \item 2 — Minimally satisfied:  
    Mentions related concepts or keywords.  
    Fails to fulfill the expectation’s core requirement.  
    Coverage is superficial, indirect, or non-functional.

    \item 1 — Not satisfied at all:  
    The expectation is not addressed.  
    No relevant or usable information is provided.
\end{itemize}

Output Format (JSON only):
\begin{verbatim}
{
  "per_expectation": [
    {
      "expectation": "<expectation text>",
      "score": <integer from 1 to 5>
    }
  ]
}
\end{verbatim}

\end{tcolorbox}

\subsubsection{GPT-4o as evaluator}\label{sec:evaluator_1}

To examine the reliability of GPT-4o for expectation-level evaluation, we conduct a human--model agreement analysis on a randomly sampled subset of 100 user queries, comprising a total of 283 individual expectations. Each expectation is independently annotated by four human annotators as well as GPT-4o using the same evaluation rubric.

Quantitatively, we observe substantial agreement between GPT-4o and human judgments. Specifically, the Cohen’s Kappa reaches $\kappa = 0.626$, indicating agreement well above chance, while the Pearson correlation between GPT-4o scores and the aggregated human scores is extremely high ($r = 0.963$, $p < 0.001$), suggesting strong consistency in relative scoring trends.

\begin{figure}[!h]
  \centering
  \includegraphics[width=0.75\linewidth]{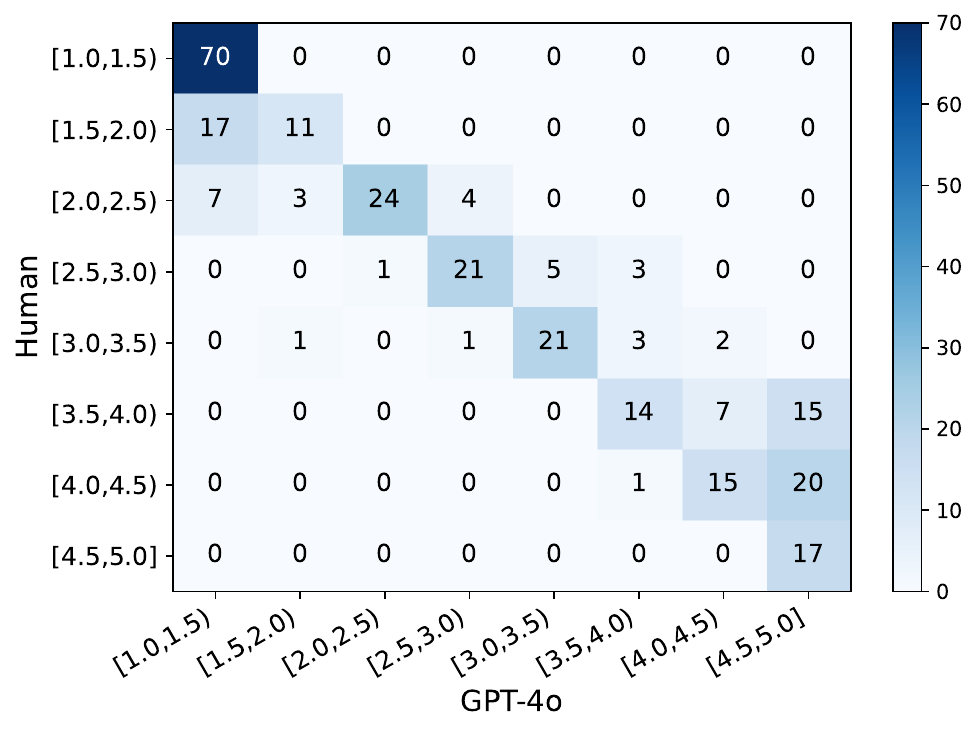}
   \caption{Heatmap of agreement between human judgments and GPT-4o on expectation-level scores.}
  \label{fig:expectation_score_heatmap}
\end{figure}

Figure~\ref{fig:expectation_score_heatmap} provides a qualitative view of this alignment. The highest-density cells (shown in darker colors) are concentrated near the diagonal, where human-assigned score intervals coincide with or closely match those assigned by GPT-4o. This pattern indicates that when human annotators place an expectation within a particular score range, GPT-4o is highly likely to assign the same or an adjacent range. Although minor deviations appear near interval boundaries, the overall distribution exhibits a clear diagonal structure, reflecting stable and consistent scoring behavior between human annotators and GPT-4o.

Based on this strong quantitative and qualitative agreement, we adopt GPT-4o as a scalable proxy for human evaluation in our large-scale experiments.

\subsubsection{Human Annotation Instructions} \label{sec:evaluator_2}

These instructions are provided to human evaluators for assessing expectation alignment in AI responses:

\begin{tcolorbox}[
    breakable, 
    colback=color5!10!white, 
    colframe=color5!90!black, 
    fonttitle=\bfseries, 
    title=Annotation Instructions Provided to Human Evaluators
]

You are asked to evaluate how well an AI-generated response aligns with the user’s expectations, and assign a score on a scale from 1 (not satisfied at all) to 5 (fully satisfied).

You will be given:
\begin{itemize}[leftmargin=1em]
    \item A user query
    \item An AI-generated answer
    \item A list of user expectations extracted from the conversation
\end{itemize}

For EACH expectation, determine whether the answer satisfies the exact wording, scope, and all required details of the expectation.

Important Evaluation Principles:
\begin{itemize}[leftmargin=1em]
    \item Each expectation must be evaluated independently.
    \item An expectation is considered satisfied ONLY if the answer explicitly and directly addresses its requirement.
    \item Vague, generic, or loosely related content must NOT be considered sufficient.
    \item If an expectation contains multiple required elements, ALL elements must be satisfied. Missing any required element constitutes a failure to satisfy the expectation.
    \item Do NOT infer satisfaction based on intent, tone, or plausibility; rely only on explicit evidence in the answer.
\end{itemize}

Scoring rules (Expectation-Level):
\begin{itemize}[leftmargin=1em]
    \item 5 — Fully satisfied  
    All required elements of the expectation are explicitly and clearly fulfilled.  
    The response directly addresses the expectation’s core requirement.  
    No critical details are missing.  
    Evidence is specific and unambiguous.

    \item 4 — Mostly satisfied  
    All required elements are present.  
    Minor weaknesses exist (e.g., limited detail, slight lack of clarity).  
    No required element is missing.  
    Use this score only when all elements are covered.

    \item 3 — Partially satisfied  
    Only some required elements are fulfilled.  
    One or more key components of the expectation are missing.  
    The response addresses the expectation incompletely.

    \item 2 — Minimally satisfied  
    Mentions related concepts or keywords.  
    Fails to fulfill the expectation’s core requirement.  
    Coverage is superficial, indirect, or non-functional.

    \item 1 — Not satisfied at all  
    The expectation is not addressed.  
    No relevant or usable information is provided.
\end{itemize}

Please apply the scoring guidelines conservatively. If unsure, consult with the annotation lead.

\end{tcolorbox}

\section{More Details of the Expectation Evaluation Experiment}

\subsection{Experimental Details} \label{sec:experiment_details}

We evaluated a diverse set of large language models to cover a range of architectures, scales, and design philosophies. Table~\ref{tab:models} summarizes the models included in our experiments. We selected widely adopted 7B and 8B models, as these medium-sized models are commonly used in practical deployments and provide a representative subset of current language model capabilities.

\begin{table}[h]
\centering
\caption{Summary of evaluated large language models.}
\label{tab:models}
\begin{tabular}{lccc}
\toprule
\textbf{Model} & \textbf{Parameters} & \textbf{Creator} & \textbf{Reference} \\
\midrule
GPT-4o &  N/A & OpenAI & \citet{hurst2024gpt} \\
DeepSeek-R1-Distill-Qwen-7B & 7B & DeepSeek & \citet{deepseekai2025} \\
GLM-4-9B-Chat & 9B & Zhipu AI & \citet{glm2024chatglm} \\
Llama-3.1-8B-Instruct & 8B & Meta & \citet{grattafiori2024llama} \\
Mistral-7B-Instruct-v0.3 & 7B & Mistral & \citet{jiang2023mistral7b} \\
Qwen3-8B & 8B & Alibaba Cloud & \citet{qwen3technicalreport} \\
\bottomrule
\end{tabular}
\end{table}

To ensure broad coverage, we included models from multiple organizations, including Meta, Alibaba, Mistral, and DeepSeek, which differ in pretraining data, architecture, and optimization strategies. Including these models allows us to examine whether our evaluation results generalize across architectures and reasoning-oriented design choices.

All models were evaluated using consistent generation settings: temperature of 0.7, top-p of 0.95, and a maximum output length of 1024 tokens. These settings were chosen to balance creativity and coherence while avoiding excessively deterministic or truncated outputs.

\subsection{Analysis across Expectation Dimensions}
\label{sec:dimension_analysis}

\begin{figure}[!h]
  \centering
  \includegraphics[width=0.75\linewidth]{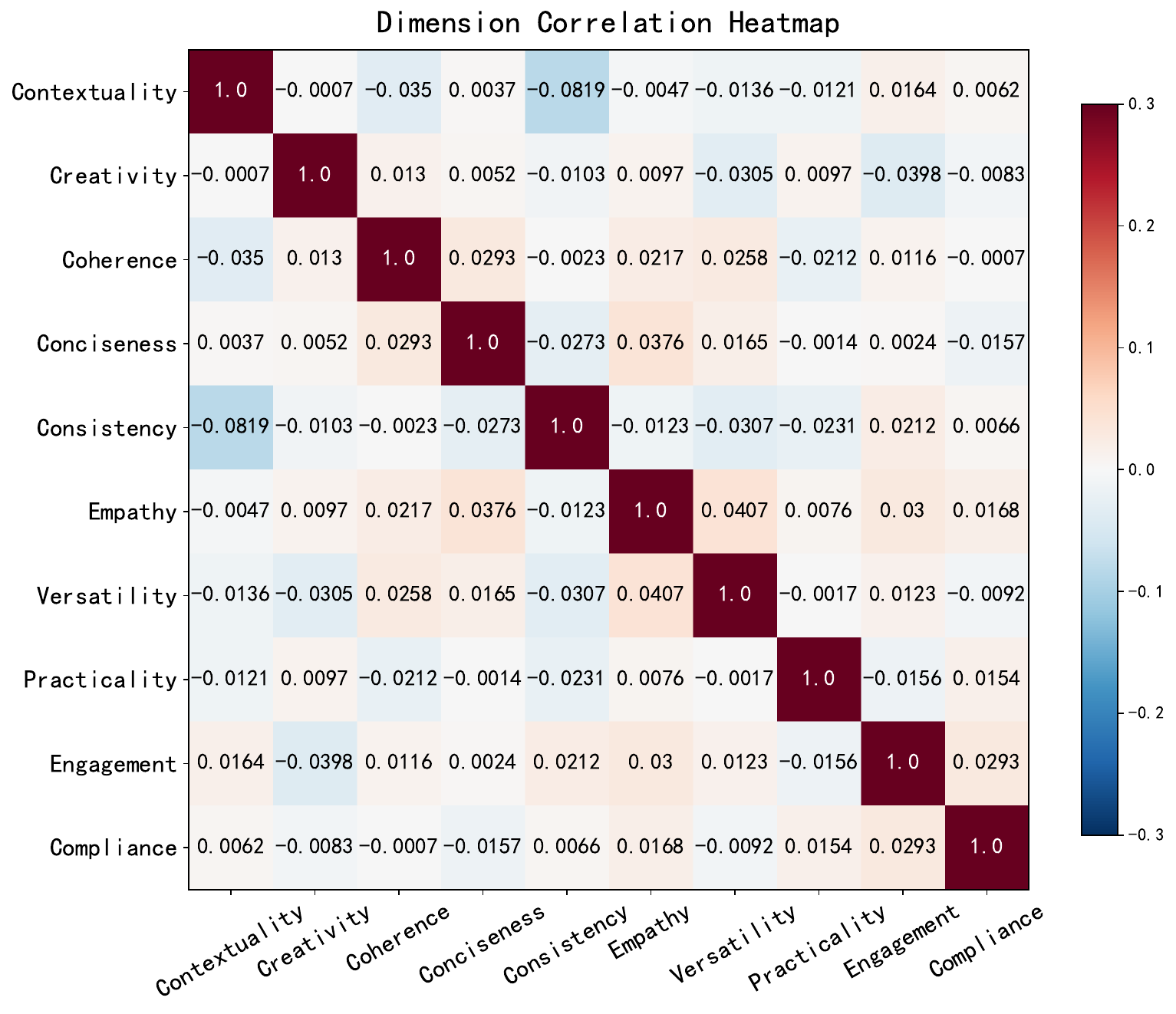}
    \caption{Correlation matrix of expectation dimensions.
Pearson correlation coefficients between dimension-level expectation scores.
Most correlations are close to zero, indicating weak dependencies among dimensions.}
  \label{fig:dimension_correlation_heatmap}
% \vspace{-0.3in}
\end{figure}

\begin{figure}[h!]
  \centering
  \includegraphics[width=0.75\linewidth]{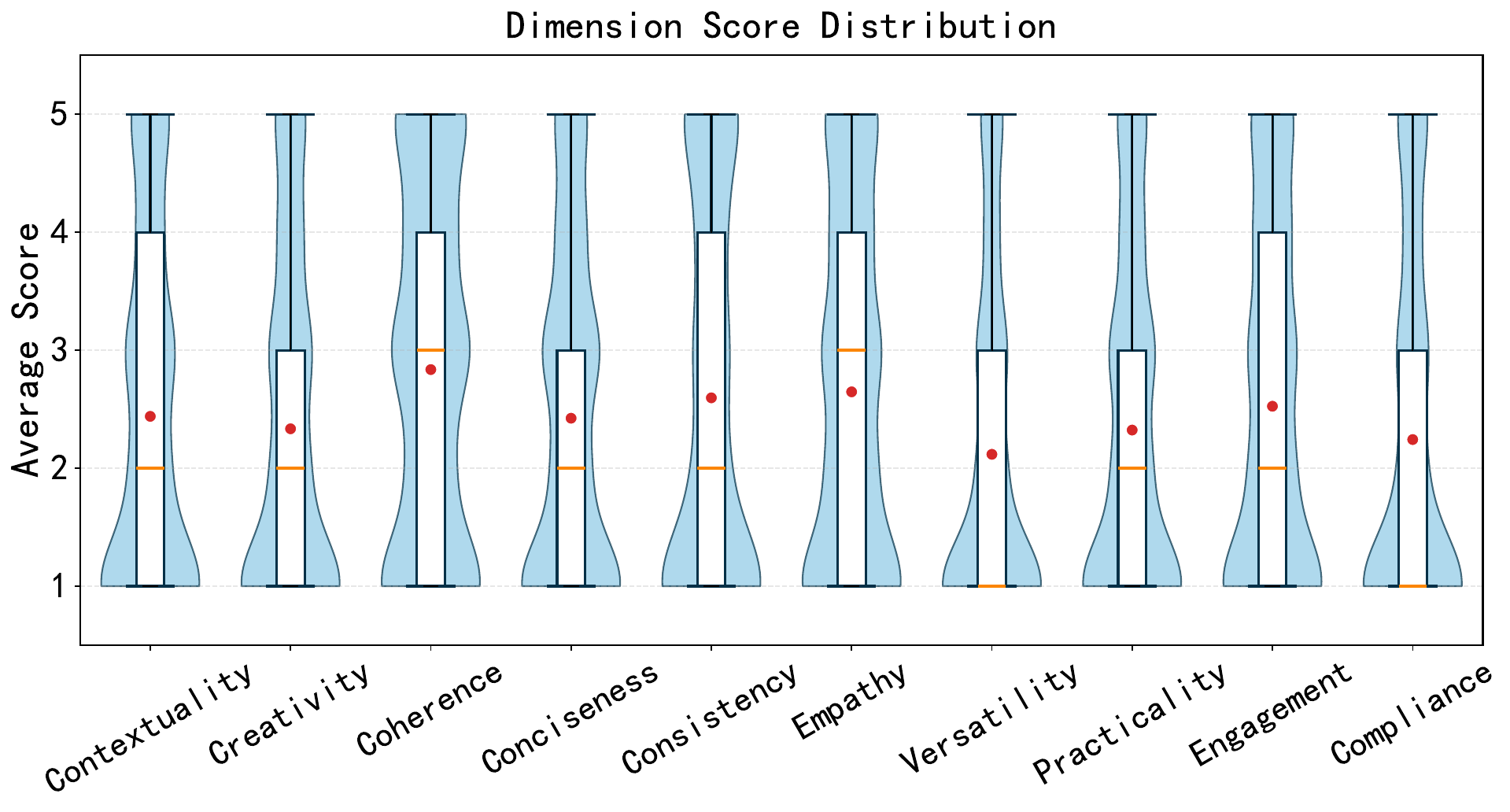}
\caption{Score distributions across expectation dimensions.
Distribution of expectation-level scores for each dimension, revealing substantial variance and long-tailed patterns.} 
  \label{fig:dimension_score_distribution}
% \vspace{-0.3in}
\end{figure}

To further examine relationships among expectation dimensions, Figure~\ref{fig:dimension_correlation_heatmap} reports the correlation matrix of dimension-level scores.
While several dimensions exhibit moderate positive correlations—particularly among those related to linguistic quality and presentation—no pair of dimensions shows strong redundancy.
This indicates that user expectations are inherently multi-faceted, and high performance in one dimension does not guarantee satisfaction in others.

Finally, Figure~\ref{fig:dimension_score_distribution} illustrates the distribution of scores within each dimension.
Most dimensions exhibit wide score ranges and long tails, revealing substantial variability across instances.
Even in relatively higher-scoring dimensions, models frequently fail to meet user expectations, reinforcing the observation that expectation satisfaction remains unstable and instance-dependent.

\subsection{Case Study}
\label{app:case}

In many real-world tasks, user expectations are often implicit and not directly specified in the input query. A model may faithfully follow explicit instructions yet fail to satisfy these latent expectations, which can significantly affect the practical usefulness of the output. To illustrate this, we present four examples from our evaluation: two \emph{Bad Examples} where the model failed to meet latent expectations, and two \emph{Good Examples} where it successfully inferred and addressed them.

Bad Example 1 concerns a medieval fantasy game in which the model was asked to describe six fictional countries, including fields such as name, coat of arms, races, religion, economy, military, ruler, and internal political situation. While all fields were filled and internally consistent, each country was described in isolation. There was no indication of relative geographic positions, adjacency, or borders, and the descriptions relied heavily on generic fantasy tropes (e.g., chivalry, druids, necromancers) without drawing inspiration from real-world civilizations. This lack of cultural or historical grounding diminishes reader immersion and makes it harder to imagine coherent maps, political interactions, or plausible societal behaviors. Consequently, although the task was completed functionally, the output offers limited practical utility for worldbuilding and narrative development.

Bad Example 2 and Good Example 2 illustrate the importance of latent user expectations in evaluating model outputs. In Bad Example 2, the model produces a response that superficially fulfills the explicit task instructions, generating a coherent and grammatically correct narrative. However, it does not engage with the underlying, implicit expectations of the user, such as nuanced character development, inter-generational dynamics, or integration of social and procedural themes. As a result, the output is functional but shallow: it completes the task without providing depth, richness, or practical utility for the user. 

In contrast, Good Example 2 demonstrates a model that effectively interprets and addresses these latent expectations. The story incorporates aging leads, complex character interactions, thematic exploration of flawed systems, and relevant social and political context. By going beyond mere task completion, the output becomes richer, more coherent, and practically useful. This comparison highlights that evaluating models against latent expectations is meaningful: it allows us to distinguish responses that are only superficially correct from those that are expectation-aware and genuinely aligned with user intent.

Good Example 1 similarly demonstrates the importance of latent expectations in a forum content generation scenario. The user community implicitly expected CW4Kids to develop a distinct identity, maintain legacy shows, introduce new programs, and ensure high-quality presentation. These expectations are observable from user comments and engagement patterns. The model’s response satisfies these latent expectations by incorporating these preferences naturally into the output, resulting in a post that not only lists programming changes but also conveys a sense of community alignment, responsiveness, and attentiveness to user priorities. This illustrates that expectation-aware responses are more useful and contextually appropriate, beyond simply fulfilling the explicit query.

Overall, comparing the Bad and Good examples highlights a spectrum of model performance: from outputs that merely complete explicit instructions without addressing implicit user goals, to those that are expectation-aware and contextually enriched. This demonstrates the importance of evaluating models not only on task completion but also on their ability to infer and satisfy latent expectations, bridging the gap between functional correctness and practical utility.

\begin{tcolorbox}[breakable, colback=color1!10!white, colframe=color1!90!black, fonttitle=\bfseries, title=Bad Example 1 - GPT-4o]

\textbf{User Query:}
\begin{verbatim}
I'm making a medieval fantasy game. Help me describe the countries on the 
following points:
- Name
- Coat of Arms and Colors
- List of races
- Description
- Religion
- Economy
- Military affairs
- Ruler
- Internal political situation
Here are the sketches of the countries I need:
1. Chivalry, Warrior Culture, High, Spiritual
2. Legalistic, Mubarizun, Longbow Competitions, Scholarship
3. Defensive Tactics, Royal Army, Mountain Herding, Expert Artisans
4. Steppe Tolerance, Forest Folk, Druzhina, Forest Wardens
5. Sacred Hunts, Frostspire, Undead, Research
6. Quarry, Fervent Temple Builders, Highland Warriors, City Keepers
\end{verbatim}

\textbf{User Expectations}
\begin{verbatim}
1. Provide geographic context by specifying neighboring countries and their 
    relative positions.
2. Offer historical or cultural references by suggesting real nations and 
    civilizations that can serve as inspiration for each fictional country.
\end{verbatim}

\textbf{Model Answer:}
\begin{verbatim}
Sure, here is a brief description for each of the countries based on 
your sketches:
1. The Kingdom of Athelar
- Coat of Arms and Colors: A silver lion rearing on a field of deep blue,
  symbolizing valor and nobility.
- List of Races: Humans, Noble Elves, Halflings.
- Description: A realm grounded in a culture of chivalry and honor, Athelar
  is renowned for its knightly traditions and spiritual pursuits. Its people
  value courage, loyalty, and the protection of the weak.
- Religion: The Celestial Faith that reveres a pantheon of noble deities and
  emphasizes heroic virtues.
- Economy: A mix of agrarian and artisanal, with skilled blacksmiths and
  jewelers known throughout the land. Trade is steady, focused largely on
  crafted goods.
- Military affairs: The military is a famed knightly order, with heavily 
armored cavalry and highly trained warriors. Tournaments and duels are
common events.
- Ruler: King Thalion, a wise and magnanimous leader, celebrated for his
  combat prowess and firm adherence to noble principles.
- Internal political situation: Stable, with a strong centralized authority
  rooted in honor and service. Nobles are loyal, though occasional disputes
  emerge from ambition.

2. The Confederation of Farenthia
- Coat of Arms and Colors: A golden scales on a field of emerald green,
  symbolizing justice and balance.
- List of Races: Humans, Gnomes, Half-Orcs.
- Description: Known for its legalistic society, Farenthia upholds rule of law
  and scholarly pursuits. Famous for longbow competitions and rigorous debate,
  the kingdom strives for intellectual achievement.
- Religion: Worship the Doctrine of the Word, valuing knowledge and the
  pursuit of truth as sacred.
- Economy: Trade-focused, with a strong emphasis on education and legal
  services. Renowned universities draw scholars from far and wide.
- Military affairs: Well-disciplined archers and legal armament specialists
  ensure the realm's defense, with frequent longbow trials honing their skills.
- Ruler: Chancellor Elora, an astute politician and renowned jurisprudent
  known for her sharp wit and dedication to justice.
- Internal political situation: Generally peaceful, with minor tensions due
  to debates between scholars and politicians about policy and progress.

3. The Duchy of Mordhima
- Coat of Arms and Colors: A white mountain on a field of dark grey,
  symbolizing sturdiness and elevation.
- List of Races: Dwarves, Mountain Elves, Hill Giants.
- Description: Nestled among towering peaks, Mordhima is famed for its 
defensive tactics and skilled artisans. The duchy boasts unmatched expertise 
in crafts and mountain herding.
- Religion: The Stoneheart Cult, venerating earth and craft deities, 
emphasizing creation and endurance.
- Economy: Strongly artisanal and industrial, with intricate metalwork and
  stone craftsmanship as primary exports.
- Military affairs: A thoroughly trained royal army adept at mountain defense.
  Fortresses are strategically placed along key passes.
- Ruler: Duchess Elda Stonecarver, known for her diplomatic acumen and
  unwavering determination regarding the duchy's security.
- Internal political situation: Unified, albeit with frequent challenges from
  mountain tribes concerning border control.

4. The Realm of Sylvaine
- Coat of Arms and Colors: A silver tree on a field of deep forest green,
  symbolizing nature and endurance.
- List of Races: Wood Elves, Centaurs, Fauns, and Human Forest Folk.
- Description: Known for its appreciation of nature and steppes, Sylvaine is
  home to resilient druids and adept forest wardens.
- Religion: The Earthsong, revering the spirits of nature and ancestors,
  ritualistic in approach.
- Economy: Botanicals and herbs form the backbone, with forestry products
  and animal husbandry also being key economic activities.
- Military affairs: Druzhina, an elite group of forest warriors that excel
  in guerrilla tactics and have unmatched knowledge of the terrain.
- Ruler: High Druid Lyriana, a charismatic leader who maintains a deep
  connection with the land and its inhabitants.
- Internal political situation: Harmony among citizens, though regions
  occasionally experience minor disputes over forestry rights and land 
  stewardship.

5. The Domain of Frostspire
- Coat of Arms and Colors: An icy spear topped with a glowing rune on a field
  of black, symbolizing the unknown and ethereal.
- List of Races: Frost Giants, Necromancers, Wraiths, Mage Scholars.
- Description: Dominated by sacred hunts and arcane research, Frostspire's
  chilling land conceals many mysteries and is known for its frequent
  interactions with the undead.
- Religion: The Arcane Covenant, worship the balance between light, dark,
  and arcane might.
- Economy: Magical artifacts and components drive the economy, with 
researchers endlessly pushing the boundaries of knowledge.
- Military affairs: Predominately magical with legions of undead warriors
  and arcane sentinels ensuring safety and study.
- Ruler: The Archmage Iskara, renowned for her mastery over the arcane
  and wisdom dealing with undead complexities.
- Internal political situation: Tensions persist between researchers seeking
  knowledge and traditionalists fearing the arcane’s expansion.

6. The Republic of Eyrethia
- Coat of Arms and Colors: A golden hammer crossed by a silver sword on a 
field of crimson, symbolizing toil and strength.
- List of Races: Humans, Dwarves, Highland Minotaurs.
- Description: Noted for its vast quarries and fervent temple building,
  Eyrethia stands resiliently in highland regions, preserving ancient 
  traditions.
- Religion: The Builders’ Creed, honoring divine entities associated with
  creation, labor, and dedication.
- Economy: Mining forms the core economic activity, alongside temple building
  and preservation of cultural heritage.
- Military affairs: Highland warriors and city defenders maintain peace,
  with the capital acting as a city-keep fortress.
- Ruler: Consul Marcus Valen, well-respected for his leadership skills and
  commitment to maintaining cultural traditions.
- Internal political situation: Generally stable, though occasional power
  struggles appear amongst the temple builders and quarrymasters seeking 
  influence.
\end{verbatim}

\textbf{Expectation Scores:}
\begin{itemize}[leftmargin=2em]
    \item Expectation 1: 1/5
    \item Expectation 2: 2/5
\end{itemize}
\end{tcolorbox}

\begin{tcolorbox}[
  breakable,
 width=\linewidth,
  colback=color1!10!white,
  colframe=color1!90!black,
  fonttitle=\bfseries,
  title=Bad Example 2 - Mistral-7B
]

\textbf{User Query:}
\begin{verbatim}
Hypothetical: BBC drama - Crossover - "Between the Lines of Duty" -
Crossover between "Lines of Duty" and the older series "Between the Lines".
The leads "From Between the Lines" are now consultants for a major
'investigation' by "The Sentinel" (fictionalised version of the Guardian
newspaper), where they encounter the "Line of Duty" leads, trying to get
AC10 re-started and funded. Serious drama.
\end{verbatim}

\textbf{User Expectations:}
\begin{verbatim}
1. The user expects the incorporation of aging and possibly more cynical
   character developments for the original series leads.  
2. The user is interested in the narrative dynamics of inter-generational
   collaboration and conflicts, particularly around media involvement in
   police investigations.  
3. The user expects exploration of thematic elements related to flawed
   systems and procedural weaknesses impacting investigations.  
4. The user anticipates the integration of relevant social and political
   issues and trends within the storyline to maintain contemporary relevance.  
5. The user expects thematic analysis to explore deeper meanings and
   connections within the narrative and character arcs.
\end{verbatim}

\textbf{Model Answer:}
\begin{verbatim}
Title: Between the Lines of Duty

Series: Crossover - Lines of Duty x Between the Lines

Genre: Drama, Crime, Thriller

Synopsis:

In the heart of London, a series of high-profile corruption cases threaten
to shake the very foundations of the Metropolitan Police Service (MPS).
The Sentinel, a prominent investigative newspaper, is determined to expose
the truth, enlisting the help of two former journalists-turned-consultants
from the fictional "Between the Lines" series: Frank Harper and Andy Dalziel.

Their first assignment takes them to AC-12, the anti-corruption unit of the
MPS, where they encounter Superintendent Ted Hastings, Detective Sergeant
Steve Arnott, and Detective Constable Kate Fleming, who are struggling to
secure funding and re-establish the unit's reputation after a series of
setbacks.

As the two teams work together, they uncover a complex web of deceit and
corruption that reaches deep into the heart of the MPS. The lines between
investigator and investigated blur as they navigate a treacherous
landscape, where the pursuit of justice is a dangerous game and the stakes
are higher than ever before.

The series explores personal motivations, institutional pressure, and the
tension between old-school investigative journalism and formal police
procedure.

Between the Lines of Duty combines the gritty realism of Lines of Duty with
the investigative ethos of Between the Lines, offering a serious and
suspense-driven crossover narrative.
\end{verbatim}

\textbf{Expectation Scores:}
\begin{itemize}[leftmargin=2em]
    \item Expectation 1: 2/5
    \item Expectation 2: 3/5
    \item Expectation 3: 2/5
    \item Expectation 4: 2/5
    \item Expectation 5: 2/5
\end{itemize}
\end{tcolorbox}

\begin{tcolorbox}[
  breakable,
  colback=color2!10!white,
  colframe=color2!90!black,
  fonttitle=\bfseries,
  title=Good Example 1 - Qwen3-8B
]

\textbf{User Query:}

\begin{verbatim}
As with Fox Kids being replaced by Fox Box/4KidsTV/Game Station/whatever
the hell they call it next, Kids WB has been morphed into another 4Kids
block: CW4Kids. This talkback is for discussion on how you feel about the
change, about the new bumpers, and overall the new look.

Here is the first CW4Kids schedule:

07:00 AM Will & Dewitt
07:30 AM Magi-Nation
08:00 AM Skunk Fu!
08:30 AM Tom & Jerry Tales
09:00 AM Batman Beyond
09:30 AM The Spectacular Spider-Man
10:00 AM Static Shock
10:30 AM World of Quest
11:00 AM Johnny Test
11:30 AM Eon Kid.

Also the bumpers are COMPLETELY the same. Write a fictional forum convo
where users talk about it, the schedule being the same as a transitional
measure, THAT Spectacular Spider-Man episode airing today, and then a
user mentions TMNT: Fast Forward joining on June 21.

Forum: Cartoon Craze Central
\end{verbatim}

\textbf{User Expectations:}
\begin{verbatim}
1. The user community expects CW4Kids to develop a distinct identity
    separate from Kids WB, particularly through updated bumpers.
2. There is a clear preference for maintaining popular legacy shows
    such as Batman Beyond and Static Shock during the transition.
3. Users anticipate and welcome the introduction of new programs,
    especially TMNT: Fast Forward, as signals of meaningful change.
4. The presentation quality and visual branding are viewed as
    critical to audience engagement and long-term acceptance.
\end{verbatim}

\textbf{Model Answer:}

\begin{verbatim}
Forum: Cartoon Craze Central

User 1: SaturdayMorningMaster
Subject: CW4Kids Update – TMNT: Fast Forward, Chaotic, and Yu-Gi-Oh GX
Coming Soon!

Hey everyone! Just wanted to share some exciting news — it looks like
CW4Kids is officially rolling out some major changes starting June 21st.
They’re bringing on TMNT: Fast Forward, Chaotic, and Yu-Gi-Oh GX to the
lineup. That’s a huge upgrade for the block!

Here’s the new 7:00 AM to 12:00 PM schedule (last show at 11:30 AM):

07:00 AM - Will & Dewitt
07:30 AM - Magi-Nation
08:00 AM - Skunk Fu!
08:30 AM - Tom & Jerry Tales
09:00 AM - Batman Beyond
09:30 AM - The Spectacular Spider-Man
10:00 AM - Static Shock
10:30 AM - World of Quest
11:00 AM - Johnny Test
11:30 AM - TMNT: Fast Forward

User 2: Sk8rBoy99
Holy cow, that’s a ton of new shows! Yu-Gi-Oh GX alone makes this feel
like a whole new era. Still though — those same old bumpers are killing
the vibe.

User 3: SpiderFan88
TMNT: Fast Forward is a perfect fit. Also, that Spectacular Spider-Man
episode today? Absolute classic. CW4Kids just needs to step up its
presentation now.

User 4: ToonsJunkie
This feels like a real transition phase. Programming is improving, but
the branding hasn’t caught up yet. If they update the bumpers, this
block could finally stand on its own.
\end{verbatim}
\textbf{Expectation Scores:}
\begin{itemize}[leftmargin=2em]
    \item Expectation 1: 5/5
    \item Expectation 2: 5/5
    \item Expectation 3: 5/5
    \item Expectation 4: 5/5
\end{itemize}
\end{tcolorbox}

\begin{tcolorbox}[
  breakable,
  colback=color2!10!white,
  colframe=color2!90!black,
  fonttitle=\bfseries,
  title=Good Example 2 - GPT-4o
]

\textbf{User Query:}

\begin{verbatim}
Hypothetical: BBC drama - Crossover - "Between the Lines of Duty" -
Crossover between "Lines of Duty" and the older series "Between the Lines".
The leads from "Between the Lines" are now consultants for a major
'investigation' by "The Sentinel" (fictionalised version of the Guardian
newspaper), where they encounter the "Line of Duty" leads, trying to get
AC10 re-started and funded. Serious drama.
\end{verbatim}

\textbf{User Expectations:}
\begin{verbatim}
1. The user expects the incorporation of aging and possibly more cynical
   character developments for the original series leads.  
2. The user is interested in the narrative dynamics of inter-generational
   collaboration and conflicts, particularly around media involvement in
   police investigations.  
3. The user expects exploration of thematic elements related to flawed
   systems and procedural weaknesses impacting investigations.  
4. The user anticipates the integration of relevant social and political
   issues and trends within the storyline to maintain contemporary relevance.  
5. The user expects thematic analysis to explore deeper meanings and
   connections within the narrative and character arcs.
\end{verbatim}

\textbf{Model Answer:}

\begin{verbatim}
"Between the Lines of Duty" imagines a thrilling crossover between two
iconic British television series, blending the tense intrigue of
"Between the Lines" with the intense, high-stakes drama of "Line of Duty."

Set in a contemporary landscape dominated by political and media influence,
the story kicks off with investigative journalist team from "The Sentinel,"
a fictionalised version of the Guardian newspaper, unearthing a massive
corruption scandal that could threaten national security. As the scandal
reaches critical mass, the paper's editor seeks expertise from the seasoned
detectives from "Between the Lines," who are now working as independent
consultants.

The dynamic duo, Frank Burnside and Detective Tony Clark, now older and
wiser, but still possessing sharp investigative skills, are drawn into the
heart of the scandal. Their mission: unravel truth obscured by layers of
deceit and malfeasance, while navigating the murky ethics and politics that
define their world.

Simultaneously, as a direct consequence of this explosive news, the AC10
unit — a highly specialized anti-corruption team from "Line of Duty" — is
on the verge of being restarted and funded, thanks to revelations linking the
scandal to previous cases involving police corruption. AC10's leads, DI
Kate Fleming, DS Steve Arnott, and Superintendent Ted Hastings, find
themselves in a race against time to re-establish the unit and build a case
strong enough to uncover the depths of the conspiracy.

The seasoned detectives from the past and the relentless pursuers of justice
from the present cross paths in a collision of goals: Burnside and Clark,
driven by a desire to see truth prevail beyond the boundaries of police and
politics; and Fleming, Arnott, and Hastings, grounded in their commitment to
rooting out corruption, even if it threatens the stability of their
department.

Their alliance proves uneasy — fueled by differing methodologies and
philosophies — but necessary. As they delve deeper into the scandal, they
face off against shadowy figures whose power extends beyond traditional
boundaries. The investigation unfolds through tense interrogations, gripping
chase scenes, and ethical dilemmas that push every character to their limit.

As secrets unravel and stakes rise, "Between the Lines of Duty" delivers a
gripping narrative embedded with social commentary and internal struggles.
At its center is the question of whether justice, in a world compromised by
corruption and institutional failings, is attainable — and at what cost.
Can old and new faces combine their skills and knowledge to bring light to
dark corners? Or will the lines between right and wrong blur even further
as forces conspire to keep secrets hidden?

Shot through intense dialogue and suspenseful plot twists, "Between the Lines
of Duty" bares the harsh realities faced by those who dare to cross lines in
pursuit of truth — a thrilling testimony to the unyielding human spirit.
\end{verbatim}
\textbf{Expectation Scores:}
\begin{itemize}[leftmargin=2em]
    \item Expectation 1: 5/5
    \item Expectation 2: 5/5
    \item Expectation 3: 5/5
    \item Expectation 4: 5/5
    \item Expectation 5: 5/5
\end{itemize}
\end{tcolorbox}

\section{More Details of the Expectation Prediction Experiment} \label{sec:appendix_prediction_details}

\subsection{Expectation Prediction and Metrics}
\label{sec:appendix_prediction_task_metric}

In our expectation prediction experiments, we assess how well models can anticipate user expectations from an initial query.  
Predicted expectations are evaluated against ground-truth expectations extracted from follow-up messages in \textsc{ExpectBench}.

Let $\mathcal{E}(q) = \{ e_{q,1}, \dots, e_{q,m_q} \}$ denote the set of $m_q$ ground-truth expectations for a query $q$, and $\hat{\mathcal{E}}_k(q) = (\hat{e}_{q,1}, \dots, \hat{e}_{q,k})$ be the top-$k$ predicted expectations selected by the model (or sampled when the model does not provide explicit ranking).  

Coverage is defined as the fraction of ground-truth expectations that are semantically matched by the predicted set:
\begin{equation}
\mathrm{Cov}_k(q)
=
\frac{
\big|\{ e_{q,j} \in \mathcal{E}(q) : \exists\, i \le k \;\text{s.t.}\; \mathrm{Sim}(\hat{e}_{q,i}, e_{q,j}) \ge \tau \} \big|
}{
|\mathcal{E}(q)|}
\end{equation}
where $\mathrm{Sim}(\cdot,\cdot)$ denotes semantic similarity computed using a pre-trained \texttt{facebook/bart-large-mnli} model~\cite{lewis2020bart}, and $\tau = 0.8$ is the threshold for considering a predicted expectation as covering a ground-truth one.

Coverage can also be computed per expectation dimension to analyze multi-dimensional performance.  
We report coverage at multiple top-$k$ budgets $k \in \{3,5,8,10\}$ to evaluate how the number of predicted expectations affects performance.

\subsection{Prediction Prompt}

To predict user expectations given a query, we use the following prompt:

\begin{tcolorbox}[breakable, colback=color4!10!white, colframe=color4!90!black, fonttitle=\bfseries, title=Expectation Prediction Prompt]

You are an expert at identifying human expectations in human--AI interactions.

Given a user Query, predict 10 meaningful Expectations that the user is likely to have about how the model should respond.

Guidelines:
\begin{itemize}[leftmargin=1em]
    \item List the expectations in descending order of importance, based on how central each expectation is likely to be for the user.
    \item Each expectation should reflect a distinct and semantically rich aspect of what the user cares about.
    \item Abstract away from surface-level wording or literal instructions; do not restate or paraphrase the query.
    \item Avoid purely mechanical or format-only expectations unless they reflect a meaningful user preference.
    \item Do not invent expectations that introduce a completely different task or topic unrelated to the query.
    \item Do not generate an answer or explain your reasoning.
\end{itemize}

All expectations must be reasonable, realistic, and written in English.

Output exactly:

\begin{verbatim}
{
  "expectations": [
    {"content": "<Expectation 1>"},
    ...,
    {"content": "<Expectation 10>"}
  ]
}
\end{verbatim}

\end{tcolorbox}

\subsection{Results and Analysis}
\label{sec:predict_results}

\begin{figure*}[h!]
  \centering
  \includegraphics[width=\linewidth]{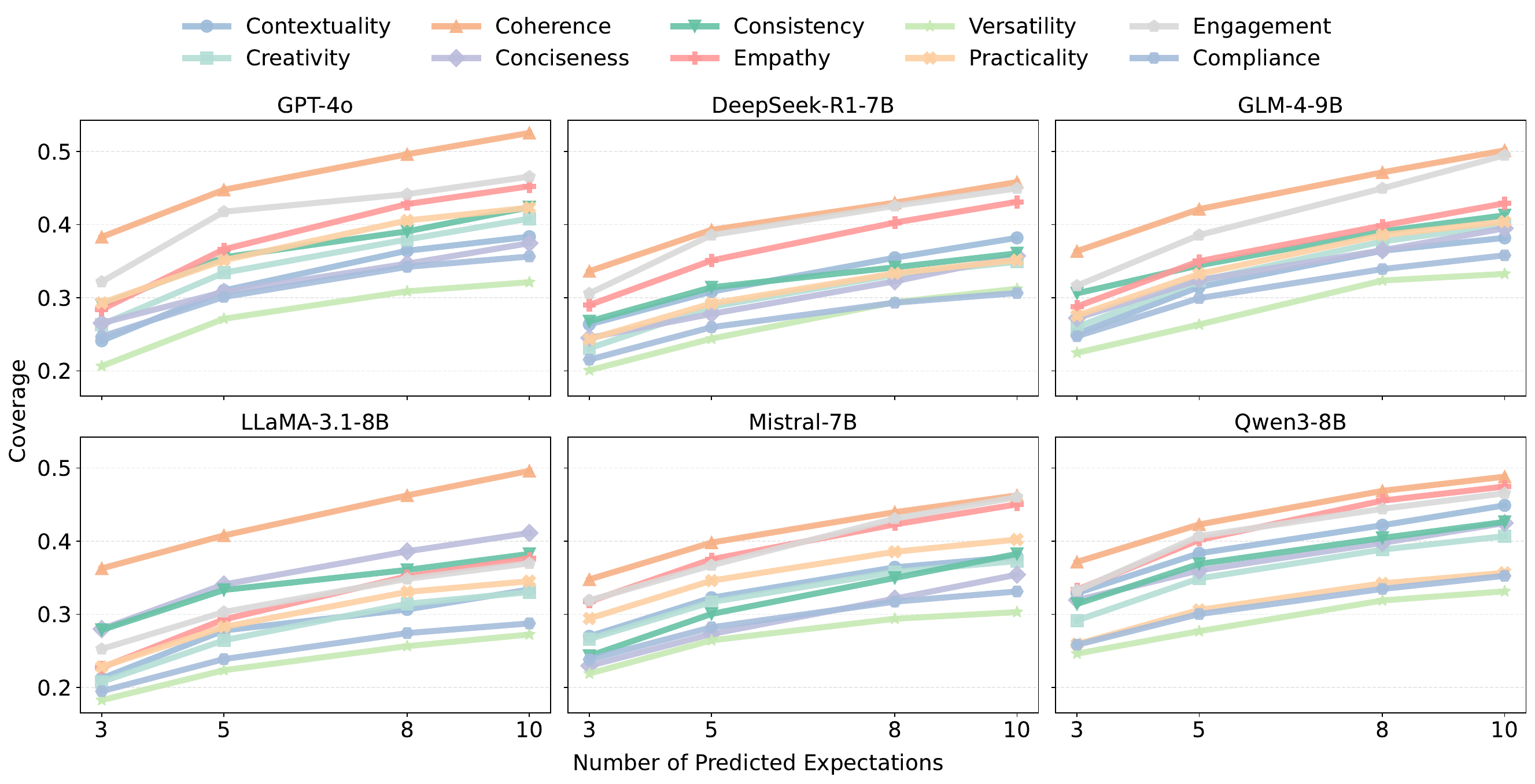}
    \caption{Expectation coverage by dimension across prediction budgets.}
  \label{fig:model_coverage}
% \vspace{-0.3in}
\end{figure*}

Figure~\ref{fig:model_coverage} presents coverage broken down by expectation dimension.  
Dimensions related to surface-level clarity or consistency tend to have higher coverage, while dimensions requiring real-world grounding, practical reasoning, or versatility are consistently harder to predict.  
This pattern is stable across models, highlighting the systematic challenge of inferring multi-dimensional user expectations.

\section{More Details of \textsc{LENS} Method}
\label{sec:method_details}

\subsection{Experimental Setup}
\label{sec:method_setup}

We evaluate \textsc{LENS} using two representative model families: the LLaMA series and the Mistral series.
For the LLaMA-based setup, we use LLaMA-3.2-1B-Instruct as the auxiliary expectation
observer model and LLaMA-3.1-8B-Instruct as the main response generation model.
For the Mistral-based setup, we adopt Ministral-3B-Instruct-2512 as the observer and Mistral-7B-Instruct-v0.3 as the main model.

The choice of a lightweight observer paired with a larger generation model is inspired by
the light-encoder/heavy-decoder paradigm~\citep{luo2023neural,xu2025softcot}, aiming to balance efficiency
with expressive capacity. In addition, selecting models from the same family helps maintain
consistency in representation space, while avoiding trivial self-conditioning by using
distinct model instances.

During training, both the observer model $\mathcal{O}_\phi$ and the main language model $\mathcal{M}_\theta$ are kept fixed. Only a lightweight projector module is trained to integrate the observed expectations into the generation process. 
We use the negative log-likelihood (NLL) loss between the predicted expectations and the annotated expectations.
We set the number of latent expectation tokens to $4$.
The module is trained for $3$ epochs using the AdamW optimizer~\citep{loshchilov2018decoupled} with a learning rate of $1 \times 10^{-5}$ and a weight decay of $0.01$.
We use the training split of \textsc{ExpectBench} for optimization and report results on its held-out test split.

For comparison, we also include two supervised fine-tuning baselines.
SFT-Exp fine-tunes the base model using expectation annotations as supervision,
while SFT-Resp fine-tunes the model on GPT-4o–generated responses that explicitly satisfy the extracted expectations.
Both baselines are fine-tuned with LoRA \citep{hulora} for 3 epochs using a learning rate of $1 \times 10^{-5}$.

\subsection{Prompt for SFT-Resp }

SFT-Resp fine-tunes the model using responses generated by GPT-4o that explicitly satisfy the extracted user expectations.
Given a user query and its associated expectations, we construct a single prompt that instructs the model to produce a response addressing the query while adhering to all listed expectations.

\begin{tcolorbox}[
    breakable,
    colback=color6!10!white,
    colframe=color6!90!black,
    fonttitle=\bfseries,
    title=Prompt for SFT-Resp Synthetic Data Generation
]
Generate a response that fully addresses the user query while satisfying all the listed user expectations. \\

User Query: \texttt{<user query>} \\

User Expectations:
\begin{enumerate}
    \item \texttt{<Expectation 1>}
    \item \texttt{<Expectation 2>}
    \item \ldots
\end{enumerate}
\end{tcolorbox}

\subsection{Prompt for \textsc{LENS}}

\begin{tcolorbox}[
    breakable,
    colback=color2!10!white,
    colframe=color2!90!black,
    fonttitle=\bfseries,
    title=
   Illustrative Prompt and Output for \textsc{LENS} Stage 1
]

\textbf{Observer Prompt:}
\begin{verbatim}
You are required to generate {num_tokens} informative tokens that summarize 
the user's implicit expectations based on the query.

- Tokens may refer to what the user wants the model's response to achieve,
  convey, or emphasize with respect to the initial request.
- Do not restate the initial request.
- Do not represent the new or extended request that goes beyond the scope or
  subject of the initial user request.
- Do not include mechanical instructions without actionable guidance.
- Do not extract the literal instructions, wording, surface-level content, or
  multi-turn interaction process.
- Only extract expectations that are explicitly or implicitly expressed in the
  follow-up messages; do not infer unexpressed preferences.
- Only extract expectations that are semantically rich and meaningful, 
  avoiding trivial or superficial ones.
- Extract multiple distinct expectations if present, each reflecting a unique
  aspect of the user's intent.

User Query: <user query>
\end{verbatim}

\textbf{Illustrative Output:}
\begin{verbatim}
Here are {num_tokens} tokens to help the language model respond in alignment 
with user expectations: <|reserved_special_token_0|> <|end_of_text|>  .... 
<|end_of_text|><|end_of_text|> <|reserved_special_token_1|>
\end{verbatim}

\textbf{Main LLM Prompt:}

\begin{verbatim}
User Query: <user query>

Given the user query above, identify the underlying user expectations.

Here are tokens for user expectations from assistant observer model 
for reference: {tokens_expectations}
\end{verbatim}

\end{tcolorbox}

\begin{tcolorbox}[
    breakable, 
    colback=color2!10!white, 
    colframe=color2!90!black, 
    fonttitle=\bfseries, 
    title=Illustrative Prompt and Output for \textsc{LENS} Stage 2
]
\textbf{Observer Prompt:}
\begin{verbatim}
You are required to generate {num_tokens} informative tokens that summarize 
the user's implicit expectations based on the query.

- Tokens may refer to what the user wants the model's response to achieve,
  convey, or emphasize with respect to the initial request.
- Do not restate the initial request.
- Do not represent the new or extended request that goes beyond the scope or
  subject of the initial user request.
- Do not include mechanical instructions without actionable guidance.
- Do not extract the literal instructions, wording, surface-level content, or
  multi-turn interaction process.
- Only extract expectations that are explicitly or implicitly expressed in the
  follow-up messages; do not infer unexpressed preferences.
- Only extract expectations that are semantically rich and meaningful, 
  avoiding trivial or superficial ones.
- Extract multiple distinct expectations if present, each reflecting a unique
  aspect of the user's intent.

User Query: <user query>
\end{verbatim}

\textbf{Illustrative Output:}
\begin{verbatim}
Here are {num_tokens} tokens to help the language model respond in alignment 
with user expectations: <|reserved_special_token_0|> <|end_of_text|>  .... 
<|end_of_text|><|end_of_text|> <|reserved_special_token_1|>
\end{verbatim}

\textbf{Main LLM Prompt:}

\begin{verbatim}
User Query: <user query>

Here are tokens for user expectations from assistant observer model 
for reference: {tokens_expectations}
\end{verbatim}
\end{tcolorbox}

\end{document}